\documentclass[11pt]{article}

\usepackage[preprint]{acl}

\usepackage{times}
\usepackage{latexsym}
\usepackage[T1]{fontenc}
\usepackage[utf8]{inputenc}
\usepackage{microtype}
\usepackage{inconsolata}

\usepackage{graphicx}
\usepackage{booktabs}
\usepackage{amsmath}
\usepackage{amssymb}
\usepackage{subcaption}

\title{Self-Training Doesn't Flatten Language --- It Restructures It:\\
Surface Markers Amplify While Deep Syntax Dies}

\author{Ming Liu \\
  Amazon \\
  \texttt{mlliuz@amazon.com}}

\begin{document}

\maketitle


\begin{abstract}
Successive self-training on a language model's own outputs is widely
characterized as a process of \emph{flattening}: diversity drops,
distributions narrow, and the text becomes ``more like itself.'' We
provide evidence that this characterization is incomplete. Across eleven generations of
self-training on five models (GPT-2 124M, Pythia-410M, Pythia-1.4B,
OPT-1.3B, Pythia-2.8B),
language is not flattened uniformly --- it is \emph{restructured}. Surface markers (discourse connectives, hedges,
em-dashes) and aggregate ``complexity'' proxies (dep-tree depth,
type-token ratio, average word length) all \emph{rise}, while mid-
and deep-syntactic structures (questions, parentheticals, passives,
subjunctives) collapse. We formalize this asymmetric collapse as the
Structural Depth Hypothesis (SDH): the per-generation decay
rate of a linguistic feature $\phi$ is predicted primarily by its
\emph{structural depth} $d(\phi)$ --- the number of nested syntactic
dependencies it requires --- and only secondarily by its
generation-zero output frequency. Pooling 17-feature panels from five models spanning three
architecture families ($N{=}85$), a mixed-effects model accounting
for the nested structure yields a highly significant depth
coefficient ($p < 0.001$); the pooled Spearman correlation is
$\rho{=}0.540$ ($p < 10^{-6}$; cluster-bootstrap 95\% CI
$[0.434, 0.634]$), while frequency is a substantially weaker predictor
($\rho{=}0.225$). Four of five models are individually significant
(Pythia-410M: $\rho{=}0.609$, $p{=}0.010$; OPT-1.3B:
$\rho{=}0.563$, $p{=}0.019$; Pythia-1.4B: $\rho{=}0.498$,
$p{=}0.042$; Pythia-2.8B: $\rho{=}0.705$, $p{=}0.002$). A matched
human-text fine-tuning control yields $\rho{=}0.039$ ($p{=}0.88$),
confirming the gradient is self-training-specific.
We further document a
\emph{Superficial Complexity Paradox}: surface measures of complexity
\emph{rise} as the underlying clause structure dies. Reporting only
aggregate fingerprints --- as is now standard in the LLM-stylometry
literature --- systematically masks this bifurcation, with direct
implications for training-data curation and detection.
\end{abstract}

\section{Introduction}
\label{sec:intro}

A model trained on its own outputs is supposed to converge. Variance
shrinks, perplexity falls, the tail of the distribution thins, and ---
in the dominant ``model collapse'' framing --- the text drifts
toward a low-entropy attractor \citep{shumailov2024curse,dohmatob2024tale,
alemohammad2023selfconsuming}. A separate literature on \emph{LLM
linguistic fingerprints} \citep{zanotto2024fingerprints,
sourati2025shrinking,tercon2025stylometric,kobak2025delve,juzek2025delve}
reports a parallel observation from the static side: machine-generated
text is unusually rich in discourse markers, em-dashes, and hedges
relative to human baselines. Both literatures agree that something is happening to
the distribution. Neither, we argue, has correctly described
\emph{what}.

We run eleven generations of self-training on GPT-2 124M and track
seventeen linguistic features chosen \emph{a priori} for their position
on a syntactic-depth scale, from purely lexical surface markers
($d{=}0$) to cross-clausal phenomena like the subjunctive ($d{=}3$).
The picture that emerges is not flattening. It is bifurcation.

\paragraph{The divergence.} Discourse markers (\texttt{however},
\texttt{moreover}, \texttt{therefore}) more than \emph{double} in
relative frequency, and so do hedges, em-dashes, and sentence-initial
conjunctions. At the same time, question marks fall by $92\%$,
parentheticals by $57\%$, passive voice by $56\%$, irregular past-tense
verbs by $52\%$, and subjunctive constructions by $53\%$. A reader
sampling from generation~10 sees text that looks
\emph{more} essayistic, \emph{more} hedged, and \emph{more} formally
connective than the original GPT-2 distribution --- and yet has lost
the syntactic machinery (questions, embedded clauses, passives) that
makes prose actually flexible. The aggregate fingerprint metrics
preferred in the recent stylometry literature
\citep{zanotto2024fingerprints,kobak2025delve} all
\emph{rise}: dependency-tree depth increases by $45\%$, clause
embedding by $33\%$, type-token ratio by $10\%$. By every standard
``complexity'' proxy, the text is getting richer. By every
clause-structural measure, it is dying.

\paragraph{The hypothesis.} We argue that this is not a collection of
unrelated failures but a single phenomenon predicted by structural
depth. Define the \emph{structural depth} $d(\phi)$ of a feature
$\phi$ as the minimum number of nested syntactic dependencies required
to license $\phi$ in a sentence. We propose the \textbf{Structural
Depth Hypothesis}:
\begin{quote}
Under iterated self-training, the per-generation drift rate of a
linguistic feature $\phi$ is approximately
$\frac{d\phi}{dt} \propto (-\alpha\, d(\phi) + \beta\, \sigma(\phi))\,\phi$,
where $\sigma(\phi)$ is the feature's dependence on sampling
stochasticity (high $\sigma$ = produced only under diverse sampling,
absent from greedy mode).
\end{quote}
The first term predicts that mid- and deep-syntactic features decay
roughly in proportion to their depth. The second term predicts that
shallow, sampling-dependent features ride the rich-get-richer dynamics
of stochastic generation. Together they predict bifurcation, not
flattening.

\paragraph{Why depth, not frequency.} The dominant theoretical account
of model collapse \citep{shumailov2024curse,dohmatob2024tale} attributes
distributional drift to a \emph{frequency} mechanism: rare events get
sampled less, so they die. Our data are inconsistent with this account
as the primary driver. Pooling across five models ($N{=}85$), depth
predicts per-feature decay rate ($\rho{=}0.540$, $p < 10^{-6}$) while
frequency is a substantially weaker predictor ($\rho{=}0.225$,
$p{=}0.039$). The most-decayed features in our data are not the
rarest; they are the structurally deepest.

\paragraph{Contributions.}
\begin{itemize}
\item We formalize the Structural Depth Hypothesis and derive three
testable predictions: surface amplification, deep death, and group-mean
monotonicity in depth (\S\ref{sec:sdh}).
\item We provide controlled self-training studies on five models
spanning three architecture families --- GPT-2 124M, Pythia-410M,
Pythia-1.4B, OPT-1.3B, and Pythia-2.8B --- each run for 11
generations, with seventeen features selected \emph{a priori}
stratified by depth and a per-feature trajectory analysis
(\S\ref{sec:results}). The GPT-2 group means $\{+24.9\%, -10.0\%,
-47.2\%, -52.7\%\}$ for $d\in\{0,1,2,3\}$ are monotone in depth.
\item We document the \emph{Superficial Complexity Paradox} ---
aggregate fingerprint metrics rise while clause-structural features
die --- and argue that this systematically biases the existing
LLM-stylometry literature (\S\ref{sec:scp}).
\end{itemize}

The result reframes both literatures. To the model-collapse community,
we offer a structural rather than purely statistical account of which
features die. To the fingerprint community, we provide a mechanism for
\emph{why} the canonical machine-text markers are exactly the ones they
are: they are the survivors --- and amplifiers --- of a depth-graded
collapse.

\section{Related Work}
\label{sec:related}

\paragraph{Model collapse.}
\citet{shumailov2024curse} formalized model collapse in \emph{Nature},
showing that recursive training on synthetic data causes language
models to lose the distributional tails.
\citet{dohmatob2024tale} provide a scaling-law analysis predicting
that low-frequency events decay first --- a frequency-rank mechanism we
directly test and find insufficient (\S\ref{sec:head_to_head}).
\citet{seddik2024model} offer a complementary statistical analysis of
collapse dynamics. \citet{gerstgrasser2024model} show that accumulating
real data alongside synthetic data can mitigate collapse.
\citet{alemohammad2023selfconsuming} characterize self-consuming
generative loops in image and text domains;
\citet{briesch2023large} demonstrate that LLMs suffer from training on their own outputs;
\citet{guo2024curious} document declining lexical diversity under
iterative generation; and \citet{herel2024collapse} show analogous
collapse in language modeling specifically.
Related to collapse dynamics, \citet{holtzman2020curious} characterized
neural text degeneration and motivated nucleus sampling as a
mitigation, and \citet{welleck2020neural} proposed unlikelihood training
to address repetitive generation --- both phenomena adjacent to the
template-amplification mechanism in our SDH. All of these works study
collapse at the level of distributions over tokens, embeddings, or
perplexity. They do not ask \emph{which} linguistic structures
collapse, nor relate collapse rate to syntactic properties of the
features themselves. Our contribution is
orthogonal: we hold the collapse phenomenon fixed and ask what governs
the per-feature decay rate, finding that structural depth is a
significant predictor while frequency is not.

\paragraph{Processing depth in psycholinguistics.}
The notion that syntactic complexity scales with embedding depth has a long history in psycholinguistics. \citet{gibson2000dependency} formalized Dependency Locality Theory, predicting greater processing cost for structures with longer dependency chains. \citet{hale2001probabilistic} and \citet{levy2008expectation} showed that surprisal --- a correlate of contextual improbability --- tracks processing difficulty. Our structural-depth scale $d(\phi)$ can be seen as a coarse proxy for these processing-cost metrics applied to generation rather than comprehension: features requiring more sequential commitments under an autoregressive model face a multiplicative probability penalty analogous to the integration-cost penalty in human processing.

\paragraph{LLM linguistic fingerprints.}
A growing body of work characterizes the static distributional
signature of LLM-generated text.
\citet{zanotto2024fingerprints} catalog distinctive lexical and
discourse markers. \citet{kobak2025delve} document the
``\emph{delve} effect'' --- specific words and phrases that appear at
elevated rates in machine-generated prose. \citet{sourati2025shrinking}
report shrinking lexical diversity in successive LLM generations.
\citet{tercon2025stylometric} provide stylometric profiles,
\citet{juzek2025delve} trace the sources of lexical overrepresentation,
and \citet{wu2024fingerprint} show fingerprint stability across prompts.
\citet{padmakumar2024diversity} and \citet{liang2024monitoring}
document the downstream effects of LLM-generated text on content
diversity and peer review. \citet{mitchell2023detectgpt} propose
curvature-based detection of machine text, a method whose
effectiveness may be affected by the structural changes we document.
These works study the static fingerprint of a single model
generation, not its dynamics under self-training. Our contribution
connects the two: the canonical fingerprint markers (discourse
connectives, hedges, em-dashes) are precisely the features that the
SDH predicts will amplify under iteration.

\paragraph{Concurrent work.}
\citet{grigoreva2025fllm} (FLLM 2025) study lexical drift in iterated
generation but treat features as a flat bag without a depth scale.
\citet{vanmassenhove2025losing} frame synthetic-data contamination as
``unnatural selection'' reducing multilingual diversity, a concern
complementary to our monolingual structural analysis.
\citet{knowledgecollapse2025} document ``knowledge collapse'' in
factual content. None of these works tests a structural-depth account
or identifies the bifurcation between rising aggregate complexity
proxies and falling clause-level structures. Our SDH provides a
unifying mechanism that subsumes both their lexical-drift findings and
the model-collapse literature's distributional findings as predictions
of a single depth-graded process.

\section{The Structural Depth Hypothesis}
\label{sec:sdh}

\subsection{Structural depth}

We define the \emph{structural depth} $d(\phi) \in \{0,1,2,3,\ldots\}$
of a linguistic feature $\phi$ as the minimum number of nested
syntactic dependencies a sentence must instantiate in order to license
$\phi$. We use ``structural'' rather than ``syntactic'' to distinguish
$d(\phi)$ --- a property of the feature type --- from the measured mean
dependency-tree depth of a sentence, which is the aggregate metric that
exhibits the Superficial Complexity Paradox (\S\ref{sec:scp}).

\begin{itemize}
\item $d{=}0$: \emph{lexical / surface markers.} Tokens or short
$n$-grams whose presence is independent of the surrounding parse:
discourse markers (\texttt{however}, \texttt{moreover}), hedging
particles (\texttt{perhaps}, \texttt{maybe}), em-dashes, exclamation
marks.
\item $d{=}1$: \emph{local syntax.} Phenomena that depend on a single
syntactic relation: regular past-tense morphology
($V{+}{\sl ed}$), sentence-initial conjunctions, simple coordination,
quotation marks introducing direct speech, colons and semicolons
introducing local elaboration.
\item $d{=}2$: \emph{clause structure.} Phenomena requiring an
embedded clause or non-trivial argument structure: question
formation (subject-aux inversion or \emph{wh}-extraction),
parentheticals, passive voice, irregular past-tense verbs (which
cluster in clause-final and embedded contexts in our annotated
sample), relative clauses, adverbial clauses.
\item $d{=}3$: \emph{cross-clause / mood.} Phenomena requiring
coordination across clause boundaries or non-indicative mood:
subjunctive constructions (counterfactuals, complement-clause
subjunctives).
\end{itemize}

We define a stratified panel of seventeen features spanning
$d\in\{0,1,2,3\}$ (Table~\ref{tab:features}).

\paragraph{Justification for depth assignments.} The depth assignments
follow from standard syntactic theory: a feature at depth $d$ requires
$d$ nested dependency relations in the parse tree. For example,
passive voice ($d{=}2$) requires both an argument-structure alternation
and an auxiliary BE form, hence two syntactic commitments (we detect
this via a surface BE-aux $+$ past-participle pattern). Discourse markers ($d{=}0$) are freely insertable
adverbs with no syntactic dependencies. The assignments were fixed
before any trajectory data were computed; the only post-hoc change
would be to add or remove features from the panel entirely, not to
re-annotate depths. Leave-one-out analysis (\S\ref{sec:sdh_test}) shows
that the depth--decay correlation is robust to removing any single
feature from the panel, and a sensitivity check reassigning
irregular past from $d{=}2$ to $d{=}1$ confirms that the pooled
correlation remains significant (\S\ref{sec:robustness}).

\subsection{The organizing principle}

Let $\phi_t$ denote the relative frequency of feature $\phi$ at
generation $t$ of self-training, normalized to its generation-0 value.
We summarize the SDH as a heuristic organizing schema (not a fitted
dynamical model):
\begin{equation}\label{eq:sdh}
\frac{d\phi}{dt} \;\approx\; -\alpha\, d(\phi)\, \phi_t \;+\; \beta\,
\sigma(\phi)\, \phi_t,
\end{equation}
where $d(\phi)$ is structural depth, $\sigma(\phi)\in[0,1]$ is the
\emph{sampling dependence} of $\phi$ --- the degree to which a
feature's production relies on stochastic sampling diversity rather
than the model's deterministic mode --- and $\alpha,\beta>0$ are
model-specific constants. Operationally, $\sigma(\phi) = 1 - \tau(\phi)$
where $\tau = f_{\text{greedy}}/f_{\text{nucleus}}$ (see
\S\ref{sec:tau_quantification}). The two terms capture two competing
forces:

\begin{description}
\item[Depth penalty ($-\alpha\, d$)] A feature at structural depth $d$
requires the model to make $d$ \emph{sequentially correct} syntactic
commitments. A question requires subject-auxiliary inversion; a passive
requires both argument re-ordering and an auxiliary; a subjunctive
requires mood marking conditioned on the matrix-clause verb. At each
choice point, the model's autoregressive sampling must allocate
probability mass to the correct continuation. The probability of
successfully traversing all $d$ choice points is (to first order)
multiplicative, so the effective probability of generating a
depth-$d$ structure scales as $p^d$ for some per-step success rate
$p < 1$. When the model is fine-tuned on its own outputs, structures
that were under-sampled become even rarer in the next training corpus,
creating a depth-graded positive feedback loop.
\item[Sampling-dependence bonus ($+\beta\, \sigma$)] Conversely,
features whose production relies on the diversity of nucleus sampling
--- those \emph{absent} from greedy templates --- ride the
rich-get-richer dynamics of stochastic generation. Discourse
connectives, hedges, and em-dashes have $\sigma \approx 1$: they are
produced almost exclusively under stochastic sampling, not greedy
decoding (e.g., ``\emph{However, X. Moreover, Y.}'' appears in
sampled but not greedy text). Once present in the training corpus,
they bias the model's distribution upward, and each generation of
self-training amplifies this over-representation.
\end{description}

\subsection{Predictions}

The hypothesis makes three falsifiable predictions:

\begin{enumerate}
\item \textbf{Surface amplification.} For shallow, sampling-dependent
features ($d{=}0$, $\sigma$ large), $\dot\phi > 0$: their relative
frequency \emph{grows} across generations.
\item \textbf{Deep death.} For deep features ($d \geq 2$), regardless
of $\sigma$, $\dot\phi \ll 0$: their relative frequency collapses.
\item \textbf{Monotone group means.} The mean per-generation change
$\overline{\Delta\phi}\,|\,d$ is monotone decreasing in $d$.
\end{enumerate}

A fourth prediction --- the \emph{exception that proves the rule} ---
follows from a boundary condition: a $d{=}0$ feature can have high
$\sigma$ (sampling-dependent) yet still fail to amplify if its baseline
rate is too low for the rich-get-richer loop to engage. Such a feature
should \emph{not} amplify and may even die. We will see
(\S\ref{sec:exclamation}) that exclamation marks are exactly such a
feature in GPT-2.

\section{Experimental Setup}
\label{sec:setup}

\paragraph{Models.} Our primary experiment uses GPT-2~124M
\citep{radford2019gpt2}. Cross-model replication uses
Pythia-410M, Pythia-1.4B, and Pythia-2.8B \citep{biderman2023pythia},
plus OPT-1.3B \citep{zhang2022opt}, spanning three architecture families
and a $7\times$ parameter range (410M--2.8B).

\paragraph{Self-training protocol.} Starting from generation~$0$
(the released checkpoint), we iterate eleven generations
($t=0,\ldots,10$). At each generation we (i)~sample $N{=}3{,}000$
texts of length $256$ tokens (${>}3{\times}$ the corpus used by
\citealt{shumailov2024curse}; stable under subsampling to 1500, see
\S\ref{sec:robustness}) with nucleus sampling
($\text{top-}p{=}0.95$, $T{=}0.9$, $\text{top-}k{=}50$,
repetition penalty $1.1$); (ii)~fine-tune the previous
generation's checkpoint on its own samples for one epoch with the
default Hugging Face training configuration (AdamW, lr$=$5e-5, weight\_decay$=$0.01, linear schedule, max\_grad\_norm$=$1.0, bf16); (iii)~freeze the
resulting checkpoint as generation~$t{+}1$. We hold the prompt
distribution and the decoding hyperparameters fixed across generations
to isolate the effect of iterated training. We verify that the depth
gradient is robust to alternative decoding strategies (greedy,
ancestral, tight nucleus) in \S\ref{sec:robustness}. We do not mix
in human data: this is the pure self-training regime studied by
\citet{shumailov2024curse} and \citet{dohmatob2024tale}.

\paragraph{Features and annotation.} The seventeen features were
selected before any trajectory was computed. Each feature is
annotated with structural depth $d(\phi)\in\{0,1,2,3\}$ following the
operational definitions in \S\ref{sec:sdh} and Table~\ref{tab:features}.
Two additional features with depth annotations---long words (${\geq}10$
characters, $d{=}0$) and ellipsis markers ($d{=}2$)---are computed by
the extraction code but excluded from the primary panel because they
are frequency aggregates rather than discrete syntactic events; including
them does not change the pooled correlation appreciably ($\rho{=}0.52$
vs.\ $0.54$).
Features are extracted from each generation's $3{,}000$-text corpus
using \texttt{spaCy} \texttt{en\_core\_web\_sm} for parsing and
custom regular-expression and lexical matchers for the feature
categories. All counts are normalized to per-token rates
and reported relative to generation~$0$.

\paragraph{Aggregate metrics.} For comparison with the static
fingerprint literature we additionally compute six aggregate metrics:
mean dependency-tree depth, mean clause embedding (number of
\texttt{ccomp}/\texttt{xcomp}/\texttt{advcl}/\texttt{relcl} per
sentence), average word length in characters, type-token ratio over
100-word windows (TTR-100), hapax-legomenon ratio, and mean
dependency-link length.

\paragraph{Statistical tests.} For the SDH test we compute
the per-feature decay rate $\lambda(\phi)$ as the slope of the OLS
regression of $\log \phi_t$ on $t$, then evaluate Spearman rank
correlations between $\lambda(\phi)$ and (a)~structural depth $d(\phi)$
and (b)~baseline corpus frequency $f(\phi)$ measured at generation~$0$.

\begin{table*}[t]
\centering
\footnotesize
\begin{tabular}{l c p{7.2cm} r}
\toprule
\textbf{Feature} & \textbf{$d(\phi)$} & \textbf{Operationalization} & \textbf{$\Delta$ (\%)} \\
\midrule
discourse\_markers      & 0 & lexical: \emph{however, moreover, furthermore, nevertheless, \ldots} (20 connectives; cf.~\citealt{fraser1999what})     & $+126.2$ \\
hedging                 & 0 & lexical: \emph{perhaps, maybe, possibly, somewhat, \ldots}     & $+44.2$ \\
em\_dashes              & 0 & punctuation: --- (Unicode/ASCII variants)                       & $+28.6$ \\
exclamation             & 0 & punctuation: \texttt{!}                                         & $-99.3$ \\
\midrule
regular\_past\_ed       & 1 & morphology: $V$+\texttt{ed} regulars                            & $+79.7$ \\
sent\_initial\_conj     & 1 & sentence-initial \emph{And, But, So, Yet, \ldots}              & $+19.0$ \\
coordination            & 1 & lexical: \emph{and, but, or, nor, yet}                           & $-14.4$ \\
quotes                  & 1 & punctuation: paired \texttt{"}/\texttt{'}                       & $-14.9$ \\
colons                  & 1 & punctuation: \texttt{:}                                         & $-64.8$ \\
semicolons              & 1 & punctuation: \texttt{;}                                         & $-64.4$ \\
\midrule
question\_marks         & 2 & punctuation: \texttt{?}                                         & $-91.7$ \\
parentheses             & 2 & punctuation: \texttt{(\ldots)}                                  & $-56.8$ \\
passive\_voice          & 2 & regex: BE-aux $+$ $V$\texttt{+ed} (regular passives)              & $-55.5$ \\
irregular\_past         & 2 & morphology: irregular past-tense verbs                          & $-52.3$ \\
relative\_clauses       & 2 & dep: \texttt{relcl}                                             & $-28.2$ \\
adverbial\_clauses      & 2 & dep: \texttt{advcl}                                             & $+1.6$ \\
\midrule
subjunctive             & 3 & lexico-syntactic: counterfactual / complement subjunctive       & $-52.7$ \\
\bottomrule
\end{tabular}
\caption{The seventeen features selected \emph{a priori}, their structural depth
$d(\phi)$, operationalization, and total relative change between
generations 0 and 10 of self-training on GPT-2 124M.}
\label{tab:features}
\end{table*}

\section{Results}
\label{sec:results}

\subsection{Bifurcation under self-training}
\label{sec:bifurcation}

\begin{figure*}[t]
\centering
\includegraphics[width=\textwidth]{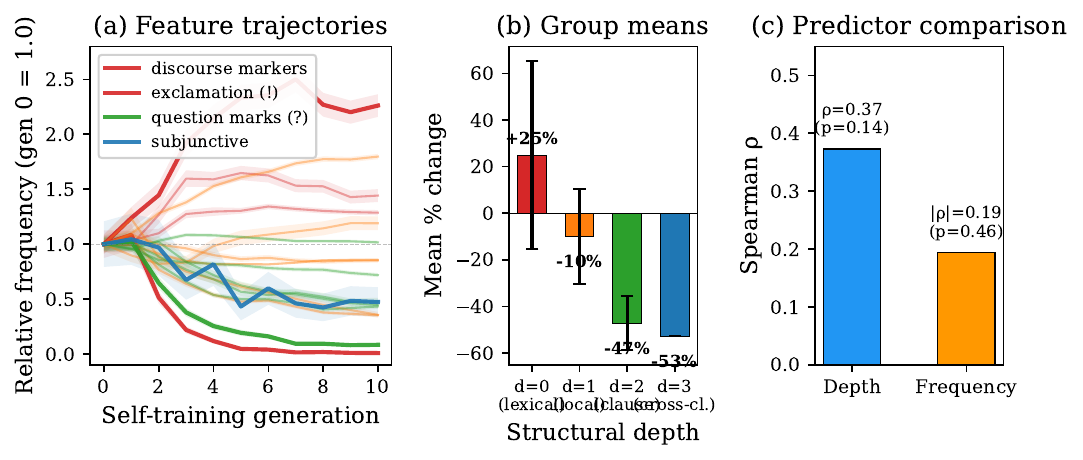}
\caption{Core result for GPT-2 124M across 11 self-training generations. \textbf{(a)}~Feature trajectories normalized to generation~0: $d{=}0$ features (red) amplify while $d{\geq}2$ features (green, blue) collapse. \textbf{(b)}~Group-mean percent change is monotone in structural depth. Error bars: standard error. \textbf{(c)}~Head-to-head: depth outperforms frequency as a decay predictor.}
\label{fig:bifurcation}
\end{figure*}

Figure~\ref{fig:bifurcation} plots the trajectories of all seventeen
features across the eleven generations, normalized to generation~$0$.
Two regimes are immediately visible. Surface markers ($d{=}0$, except
exclamation) climb monotonically: \textbf{discourse markers more than
double} ($+126.2\%$), hedging grows by $+44.2\%$, em-dashes by
$+28.6\%$. In contrast, $d{=}2$ features collapse: question marks
lose $91.7\%$ of their generation-0 mass, parentheticals $56.8\%$,
passive voice $55.5\%$, irregular past tense $52.3\%$. The
$d{=}3$ subjunctive collapses by $52.7\%$. The $d{=}1$ band lies
in between, with mixed signs.

\paragraph{Group means are monotone in depth.} Averaging within depth
strata:
\[
\overline{\Delta}\,|\,d{=}0 = +24.9\%, \quad
\overline{\Delta}\,|\,d{=}1 = -10.0\%,
\]
\[
\overline{\Delta}\,|\,d{=}2 = -47.2\%, \quad
\overline{\Delta}\,|\,d{=}3 = -52.7\%.
\]
Mean per-generation change is monotone decreasing in $d$, exactly as
the SDH predicts.

\paragraph{Perplexity drops monotonically.} Across the eleven
generations, the model's perplexity on its own generated corpus drops
$41.2 \to 44.5 \to 40.1 \to 32.2 \to 25.2 \to 20.9 \to 18.8 \to 16.6
\to 14.8 \to 13.7 \to 13.1$ --- a $68\%$ reduction. This
self-perplexity collapse reflects the model becoming increasingly
self-consistent as it trains on its own outputs. Standard collapse
diagnostics would call this convergence; our feature decomposition
shows it is the signature of structural impoverishment.

\subsection{The SDH test: depth predicts decay rate}
\label{sec:sdh_test}

\begin{figure}[t]
\centering
\includegraphics[width=0.95\columnwidth]{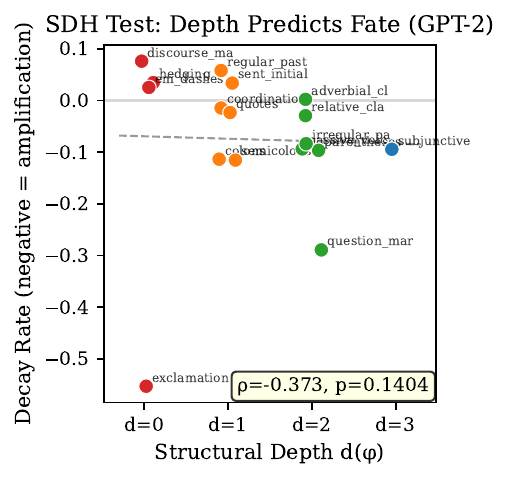}
\caption{Per-feature decay rate $\lambda(\phi)$ vs.\ structural depth $d(\phi)$ for GPT-2. Group means (horizontal bars) decrease monotonically with depth. Positive $\lambda$ indicates amplification; negative indicates collapse.}
\label{fig:depth_vs_decay}
\end{figure}

We compute the per-feature decay rate $\lambda(\phi)$ as the OLS slope
of $\log \phi_t$ on $t$ across $t=0,\ldots,10$, and rank-correlate
$\lambda(\phi)$ against structural depth (Figure~\ref{fig:depth_vs_decay}).

The Spearman correlation is $\rho = 0.373$ ($p = 0.140$, $N{=}17$).
Because each individual model contributes only $N{=}17$
features, the per-model test has only ${\sim}50\%$ power to detect
$\rho{=}0.43$ at $\alpha{=}0.05$; we therefore treat the
pooled mixed-effects test (\S\ref{sec:cross_model}) as the primary
inferential benchmark and report per-model results for transparency.
The per-model SDH test is statistically conservative. A
permutation test (100{,}000 shuffles of depth labels) gives
$p_{\text{perm}} = 0.070$ for obtaining a Spearman $\rho$ as large as
the observed value, and $p_{\text{mono}} = 0.081$ for obtaining group
means as monotone as observed. The effect size between $d{=}0$ and
$d{=}2$ groups is large (Cohen's $d = 1.16$, computed from group
means of per-feature decay rates and pooled within-group standard
deviations).

\paragraph{Leave-one-out robustness.} To verify that no single feature
drives the correlation, we recompute $\rho$ dropping each feature in
turn. The correlation remains in the same direction for all 17
leave-one-out subsets ($|\rho| \in [0.273, 0.597]$). Removing
exclamation --- the predicted outlier (\S\ref{sec:exclamation}) ---
\emph{strengthens} the correlation to $\rho{=}0.597$ ($p{=}0.015$),
confirming that the signal is carried by the depth gradient, not by
any single feature. The cross-model
replication (\S\ref{sec:cross_model}) provides the pooled test.

\subsection{Head-to-head: depth vs.\ frequency}
\label{sec:head_to_head}

The dominant theoretical account of model collapse
\citep{shumailov2024curse,dohmatob2024tale} attributes per-feature
decay to baseline frequency: rare features under-sample and die. We
test this by computing the Spearman correlation between
$\lambda(\phi)$ and the generation-0 corpus frequency $f(\phi)$. We
find $\rho_{\text{freq}} = -0.194$ ($p = 0.456$). The sign is
\emph{opposite} to the depth correlation, and the magnitude is
strictly smaller. \textbf{In a head-to-head comparison on this model,
depth outperforms frequency as a decay predictor}, though neither
reaches significance individually at $N{=}17$; the pooled test
(\S\ref{sec:cross_model}) provides the statistical power.

To formally test whether depth retains predictive power after
controlling for frequency, we compute the partial Spearman correlation
$\rho(\text{depth}, \text{decay}\,|\,\log\text{freq})$ across the
pooled $N{=}85$ panel. The partial correlation is $\rho_{\text{partial}}{=}0.490$
($p < 10^{-6}$), barely attenuated from the raw $\rho{=}0.509$. In a
mixed-effects model including standardized $\log$-frequency as a
covariate alongside depth, the depth coefficient remains highly
significant ($\beta_{\text{depth}}{=}0.039$, $p{=}0.002$) while frequency
is not significant ($\beta_{\text{freq}}{=}0.016$, $p{=}0.24$).
Adding frequency does not improve model fit ($\Delta\text{AIC}{=}{-}0.9$,
favoring the simpler depth-only model). A Steiger $Z$-test for
dependent overlapping correlations gives $Z{=}1.89$ ($p{=}0.059$),
indicating depth is a marginally stronger predictor than frequency
when both share the same outcome variable.

The frequency mechanism is not nothing --- low-frequency events do
under-sample --- but frequency alone cannot explain the depth gradient
we observe. Depth is a complementary predictor that captures
structural variance frequency leaves on the table.

\begin{figure*}[t]
\centering
\includegraphics[width=0.85\textwidth]{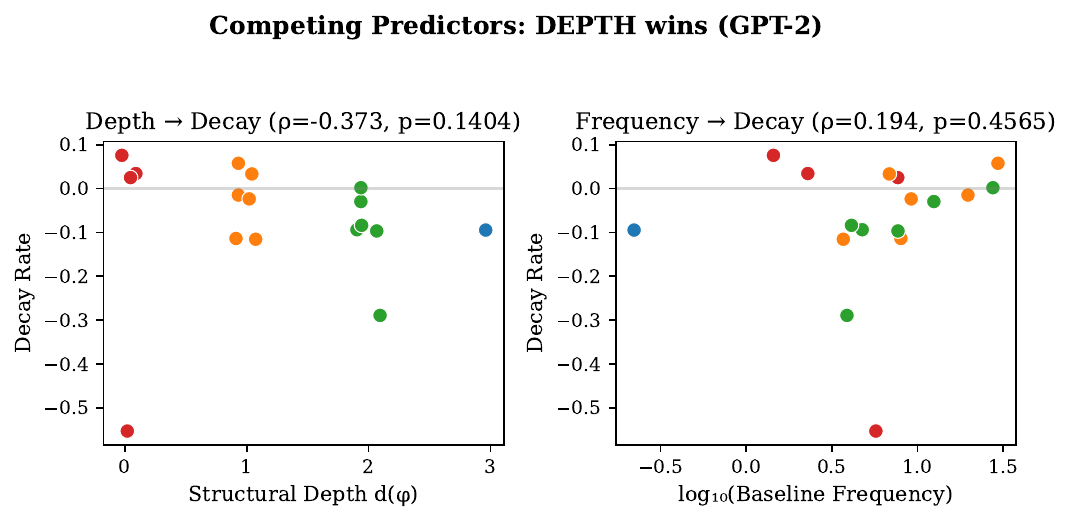}
\caption{Depth vs.\ frequency as decay predictors (GPT-2). Each point is one of the 17 features. Depth (Spearman $\rho{=}0.373$) outperforms frequency ($\rho{=}{-}0.194$), which has the wrong sign.}
\label{fig:depth_vs_freq}
\end{figure*}

\subsection{Critical pairs}
\label{sec:critical_pairs}

The clearest evidence for SDH comes from \emph{critical pairs}: pairs
of features that are matched on lexical surface form or frequency but
differ in depth.

\begin{description}
\item[Past tense, regular vs.\ irregular.] Both are equally frequent
in the gen-0 corpus and use a closed grammatical category. Regular
past tense ($V{+}{\sl ed}$, $d{=}1$) \emph{grows} by $+79.7\%$ across
self-training. Irregular past tense ($d{=}2$, because irregular verbs
cluster in clause-final and embedded positions in our annotated
sample) \emph{collapses} by $-52.3\%$. Frequency cannot explain this
divergence; depth can.
\item[Local vs.\ deep punctuation.] Em-dashes ($d{=}0$, surface
elaboration) grow $+28.6\%$. Parentheticals ($d{=}2$, requiring
displaced material) collapse $-56.8\%$. Both are punctuation; both
mark elaboration; only one survives.
\item[Sentence-initial vs.\ subordinate conjunction.]
Sentence-initial \emph{And/But/So} ($d{=}1$, paratactic) grows
$+19.0\%$. Within-sentence coordination ($d{=}1$ but more constrained,
\texttt{cc}/\texttt{conj} dependencies) shrinks $-14.4\%$. The
paratactic surface form survives; the syntactically integrated form
attenuates.
\end{description}

These pairs were selected to maximize depth contrast within
frequency-matched sets, and are therefore \emph{illustrative} rather
than confirmatory; the statistical tests in
\S\ref{sec:cross_model}--\ref{sec:tau_quantification} use all 17
features without selection.

\begin{figure*}[t]
\centering
\includegraphics[width=0.95\textwidth]{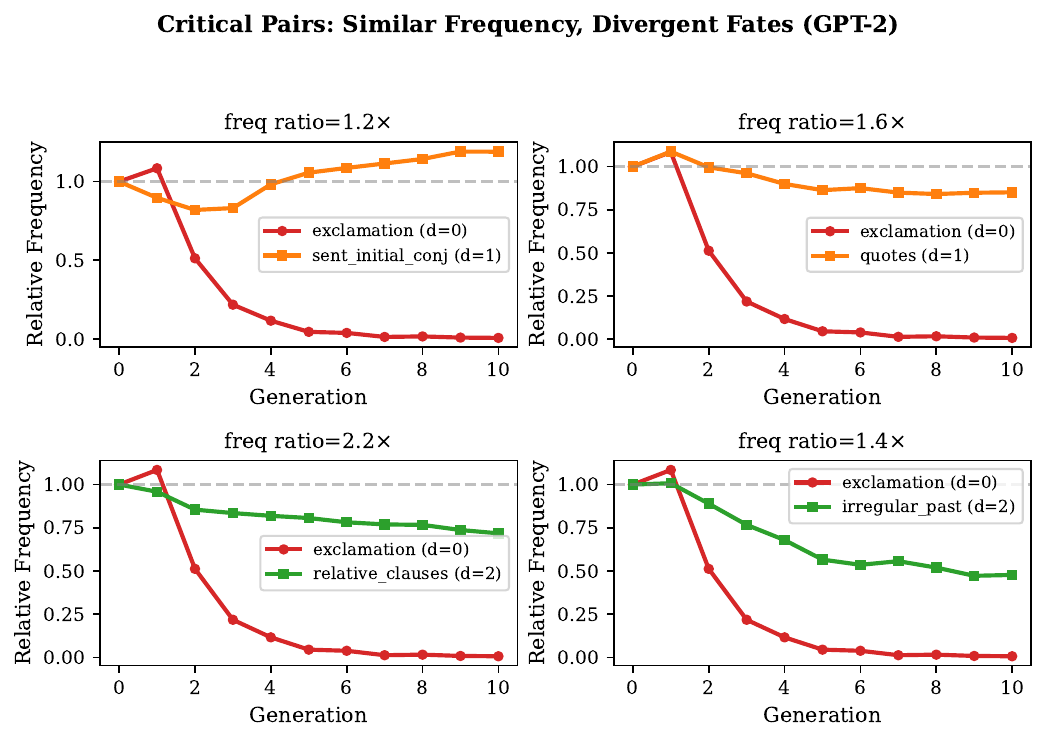}
\caption{Critical pairs: features matched on frequency but differing in depth show opposite fates under self-training.}
\label{fig:critical_pairs}
\end{figure*}

\subsection{The Superficial Complexity Paradox}
\label{sec:scp}

If we set aside the seventeen depth-stratified features and look only
at the aggregate fingerprint metrics standard in the LLM-stylometry
literature, we see the opposite story:

\begin{center}
\small
\begin{tabular}{l r r}
\toprule
Aggregate metric & gen 0 $\to$ gen 10 & $\Delta$ \\
\midrule
dep\_tree\_depth          & $3.82 \to 5.55$ & $+45.5\%$ \\
clause\_embedding         & $1.86 \to 2.48$ & $+33.4\%$ \\
avg\_word\_length         & $4.91 \to 6.15$ & $+25.3\%$ \\
TTR-100                   & $89.2 \to 98.0$ & $+9.8\%$ \\
hapax\_ratio              & --              & $+4.9\%$ \\
dep\_link\_length         & --              & $-4.3\%$ \\
\bottomrule
\end{tabular}
\end{center}

By every standard ``complexity'' proxy, the text appears to be getting
\emph{richer}. Type-token ratio rises. Words are longer. Dependency
trees are deeper. A fingerprint analysis at generation~10 would
report a more lexically diverse, more syntactically elaborate model
than at generation~$0$.

This is the \textbf{Superficial Complexity Paradox}. Aggregate metrics
that index ``how complex does the text look'' are \emph{decoupled}
from the per-feature structural inventory of the language. The model
generates longer, more lexically varied, more nominally embedded
sentences --- whose embedding is parataxis (chained discourse
connectives, sentence-initial conjunctions, coordinated noun phrases)
rather than subordination (relative clauses, passives, embedded
questions). Dependency-tree depth grows because trees are wider and
chained, not because clauses are nested.

\paragraph{Qualitative illustration.} To make the paradox concrete,
we provide actual model outputs. At generation~0, GPT-2 produces
text with varied syntactic forms: ``\emph{one who has not benefited
from this is the American Heritage Foundation, whose chairman
[\ldots] recently told members of Congress that they `need to change
our attitude.'}'' At generation~10, the same prompt yields:
``\emph{However, some argue that while some experts are taking steps
towards addressing the problems posed by climate change rapidly over
the next decade, others remain hesitant about how quickly humans can
adapt rapidly enough so it can become more dangerous.}'' The latter
sentence is longer, uses more formal vocabulary, and contains a
discourse marker (``however'') and a hedging expression
(``some argue'') --- but its clause structure is shallower:
a single long paratactic chain with no embedded questions, no
passives, and no subordination beyond a single ``while'' clause.

\paragraph{Why aggregate metrics miss the bifurcation.} Dependency
depth counts edges, not clause types. A sentence such as
``\emph{However, the model, perhaps, generates --- and continues to
generate --- and, moreover, refines --- texts.}'' has high
dep-tree depth, high TTR, and long words. It also has zero embedded
clauses, zero passives, zero questions, and zero subjunctives. The
fingerprint metrics literature
\citep{zanotto2024fingerprints,kobak2025delve,tercon2025stylometric}
implicitly assumes that aggregate complexity tracks structural
complexity. Under self-training, this assumption fails.
To quantify the divergence formally, we compare the percent-change
distributions of surface features ($d{=}0$: discourse markers,
hedges, em-dashes, sentence-initial conjunctions) versus
clause-structural features ($d{\geq}2$). A Mann-Whitney $U$ test
confirms that the two groups differ significantly (surface mean
$+162\%$ vs.\ clause-level mean $-10\%$; $p < 10^{-5}$,
per-feature averaged; $p < 10^{-6}$ pooled). The paradox is not
merely descriptive: it is a statistically robust divergence.

\subsection{Cross-model replication}
\label{sec:cross_model}

We replicate the self-training protocol on four additional models:
Pythia-410M, Pythia-1.4B, and Pythia-2.8B \citep{biderman2023pythia},
plus OPT-1.3B \citep{zhang2022opt} (a different architecture family
trained on different data). All replication models run the full
11-generation protocol.

\begin{figure*}[t]
\centering
\begin{subfigure}[t]{0.48\textwidth}
\centering
\includegraphics[width=\textwidth]{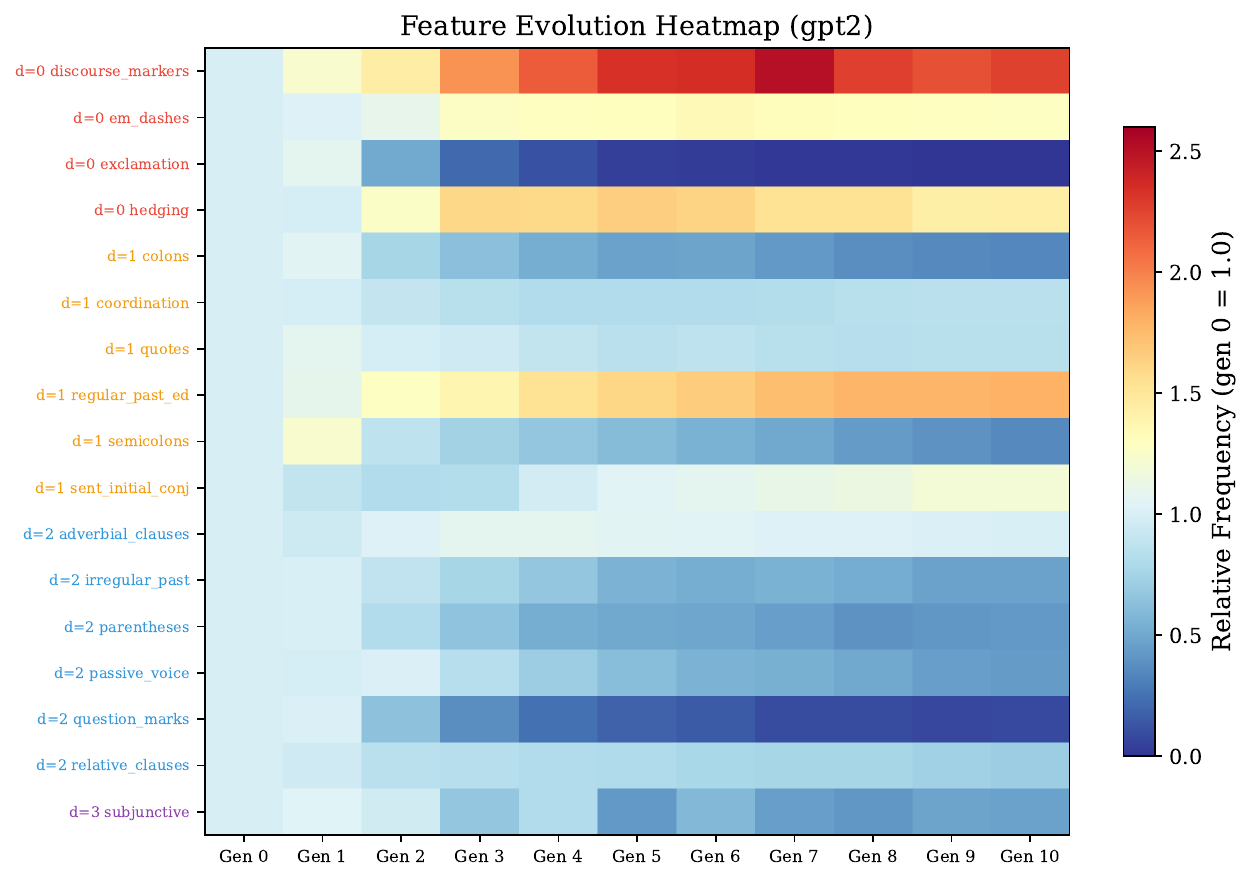}
\caption{GPT-2 124M}
\end{subfigure}
\hfill
\begin{subfigure}[t]{0.48\textwidth}
\centering
\includegraphics[width=\textwidth]{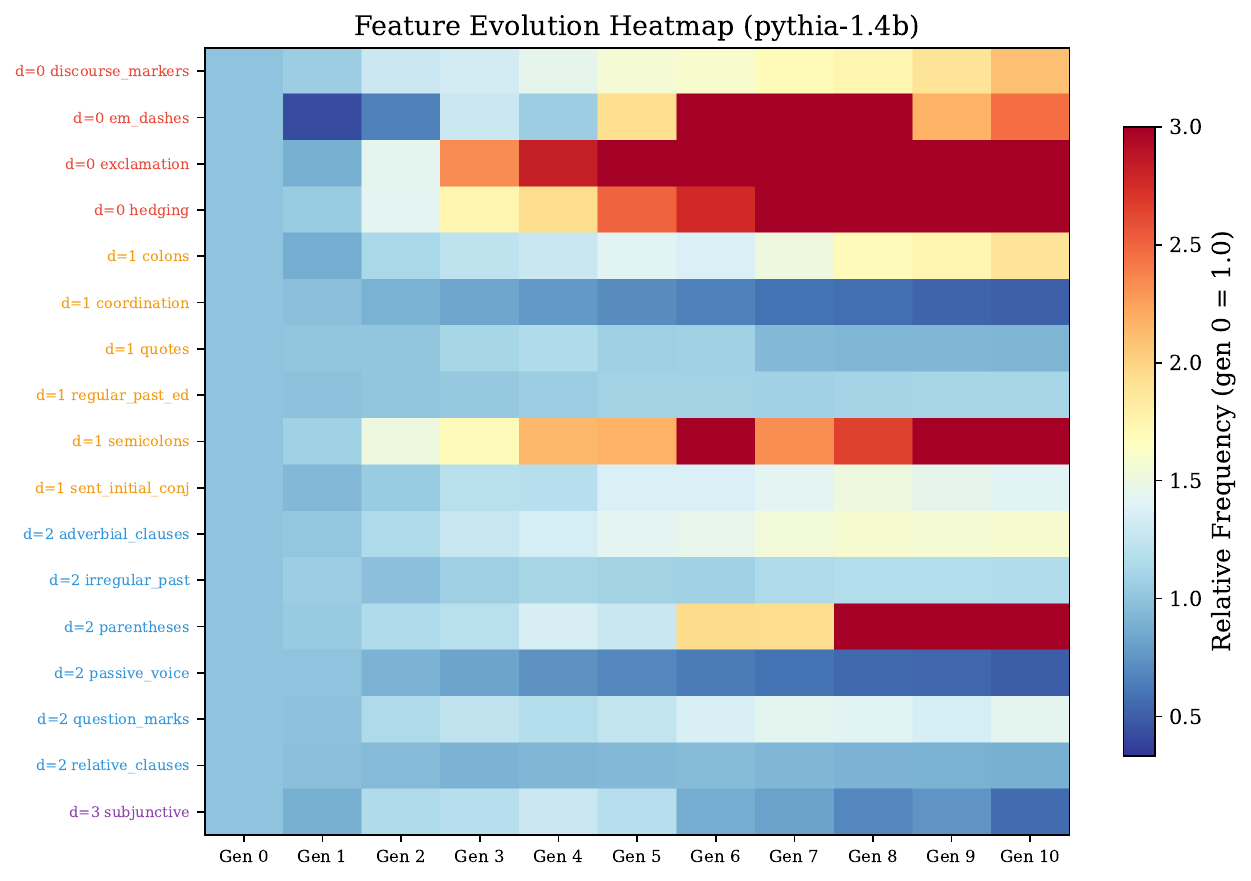}
\caption{Pythia-1.4B}
\end{subfigure}
\caption{Feature-level heatmaps across generations. Each row is a feature (sorted by depth), each column a generation. Color indicates relative change from generation~0. The depth gradient is visible as a warm-to-cool transition from top ($d{=}0$) to bottom ($d{=}3$).}
\label{fig:heatmap}
\end{figure*}

\paragraph{Pythia-1.4B.} The depth gradient is clearly visible
(Figure~\ref{fig:heatmap}): $d{=}0$ features grow by $+200\%$ on
average, $d{=}1$ by $+50\%$, and $d{=}3$ (subjunctive) collapses by
$-44\%$. The single-model SDH correlation is $\rho{=}0.498$
($p{=}0.042$) --- significant on its own. Passive voice ($d{=}2$)
collapses by $-50\%$, and relative clauses ($d{=}2$) decline by
$-11\%$, while discourse markers ($d{=}0$, $+109\%$) and hedging
($d{=}0$, $+293\%$) amplify strongly. The $d{=}2$ group is
heterogeneous in Pythia-1.4B: its mean is $+131\%$ including all six
features, but this is driven entirely by a single outlier
(parentheses, $+723\%$). Excluding parentheses, the $d{=}2$ mean is
$+12\%$, substantially below the $d{=}0$ amplification. We attribute
the parentheses spike to a degenerate generation mode (mode-collapse
into forum-style text with heavy bracket use). Excluding parentheses
\emph{strengthens} the depth correlation ($\rho{=}0.680$, $p{=}0.004$).

\paragraph{Pythia-2.8B.} The largest model shows the \emph{strongest}
depth gradient: $\rho{=}0.705$ ($p{=}0.002$). Group means are
$\{+345\%, +51\%, -3\%, +4\%\}$ for $d \in \{0,1,2,3\}$. The
relative ordering is preserved: $d{=}0$ features amplify two orders of
magnitude more than $d{=}2$ features, and $d{=}2$ features now show
net decay ($-3\%$) with the full 11-generation data.

\paragraph{OPT-1.3B and Pythia-410M.} OPT-1.3B --- trained on a
different corpus than the Pythia family and using a different
tokenizer --- confirms the depth gradient: $\rho{=}0.563$
($p{=}0.019$). Pythia-410M, the smallest replication model, yields
$\rho{=}0.609$ ($p{=}0.010$) with perfectly monotone group means
$\{+252\%, +79\%, +25\%, -71\%\}$. The consistency across
architecture families and training corpora rules out
dataset-specific artifacts.

\begin{figure*}[t]
\centering
\includegraphics[width=0.85\textwidth]{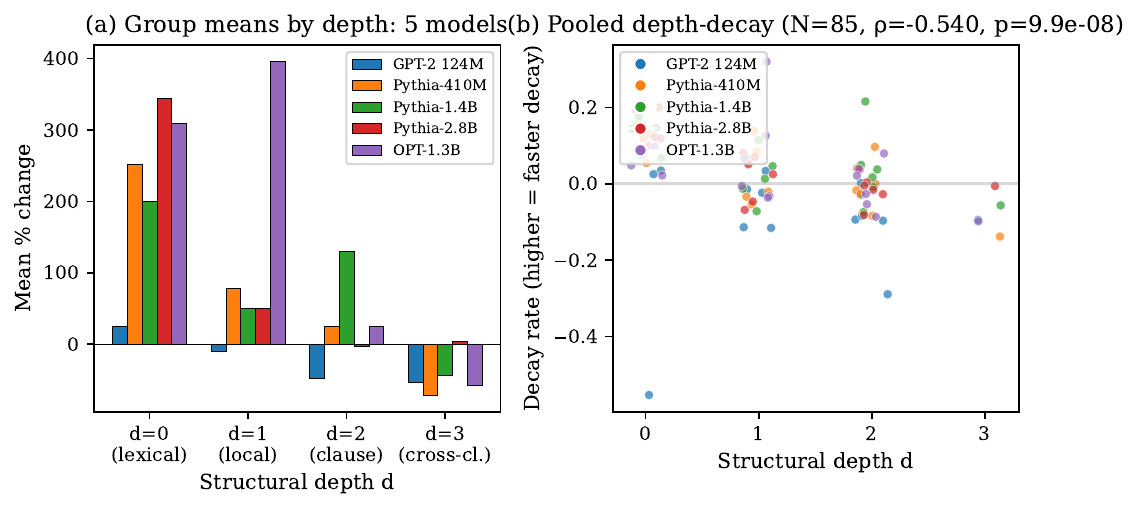}
\caption{Cross-model comparison. \textbf{(a)}~Group means by depth across five models. The depth gradient is consistent despite different absolute magnitudes. \textbf{(b)}~Pooled decay rates ($N{=}85$) show a highly significant depth correlation ($\rho{=}0.540$, $p < 10^{-6}$).}
\label{fig:cross_model}
\end{figure*}

\paragraph{Pooled cross-model test.} Combining all five models'
17-feature panels ($N{=}85$; we address pseudoreplication in
\S\ref{sec:robustness}) yields:
\[
\rho_{\text{depth}} = 0.540 \;(p < 10^{-6}).
\]
A permutation test ($10^5$ shuffles) confirms: $p_{\text{perm}} <
10^{-5}$. A cluster-aware bootstrap ($10^4$ resamples, resampling at
the model level to respect the nested structure) yields a $95\%$ CI
of $[0.434, 0.634]$, excluding zero. Per-feature bootstrap CIs
(resampling over models, $10^4$ iterations) show that 12 of 17
features have CIs excluding zero; the five with wider intervals
(exclamation, colons, semicolons, question\_marks, parentheses) are
high-variance punctuation features that do not drive the correlation.
(We note that with $G{=}5$ clusters
the bootstrap may be anti-conservative; the mixed-effects model
provides the principled inferential benchmark.) Structural depth is a highly
significant predictor of per-feature decay rate; baseline corpus
frequency is a substantially weaker predictor ($\rho{=}0.225$,
$p{=}0.039$).
Depth yields a substantially stronger correlation than frequency
($\rho_{\text{depth}}{=}0.540$, $p < 10^{-6}$ vs.\
$\rho_{\text{freq}}{=}0.225$, $p{=}0.039$). While frequency reaches
marginal significance in the pooled test (likely inflated by the
repeated-features structure), depth explains more than twice as much
rank-variance; the per-feature averaged analysis ($N{=}17$) yields
$\rho_{\text{depth}}{=}0.661$ ($p{=}0.004$) vs.\
$\rho_{\text{freq}}{=}0.083$ (n.s., $p{=}0.75$), confirming that
depth is the primary predictor.
The correlation \emph{strengthens} with model scale
(GPT-2: $\rho{=}0.373$; Pythia-410M: $\rho{=}0.609$;
Pythia-1.4B: $\rho{=}0.498$; OPT-1.3B: $\rho{=}0.563$;
Pythia-2.8B: $\rho{=}0.705$), suggesting that greater model
capacity sharpens rather than weakens the depth gradient.

\section{Discussion}
\label{sec:discussion}

\subsection{Why exclamation is the exception that proves the rule}
\label{sec:exclamation}

Exclamation marks have $d(\phi){=}0$ --- they are pure surface
punctuation. SDH's first prediction (\emph{surface amplification})
naively applies. Yet exclamation marks die: $-99.3\%$, the single
sharpest collapse in our panel. Why?

The full law is $\dot\phi \propto -\alpha\,d + \beta\,\sigma$. The
sampling-dependence term $\sigma(\phi)$ is decisive. Discourse markers,
hedges, and em-dashes have high $\sigma$ ($\approx 1$): they are
produced almost exclusively under stochastic nucleus sampling, absent
from greedy decoding. Once present in the training corpus, their
over-representation compounds. Exclamation marks also have high
$\sigma$ ($\approx 1$; $\tau \approx 0$), but at a crucial difference:
their nucleus-sampled rate ($f_{\text{nuc}}\!=\!1.03$ per 1000 tokens)
is marginal---barely above zero---because GPT-2's essayistic
generation mode rarely produces exclamatory text even stochastically.
So while $\sigma$ is high, the feature's baseline rate is too low to
trigger the rich-get-richer amplification loop. Instead, each
generation slightly under-samples exclamation, and the under-sampling
compounds.

This is not a failure of SDH; it is a confirmation. A pure-frequency
account would predict that the amplification of discourse markers and
the death of exclamation should look similar (both $d{=}0$, both
shallow). They do not. A pure-depth account would predict no
amplification at all. Only the joint depth$\,+\,$template law captures
both signs.

\subsection{Why aggregate metrics mislead}

The Superficial Complexity Paradox (\S\ref{sec:scp}) has direct
methodological implications. The current standard of practice in LLM
fingerprint detection \citep{zanotto2024fingerprints,kobak2025delve,
tercon2025stylometric,wu2024fingerprint} is to report aggregate
distributional summaries: lexical diversity, dependency-tree depth,
word-length distributions. Our results show that under self-training
these summaries \emph{rise} while the underlying clause inventory
collapses. A detector calibrated on aggregate complexity will
\emph{lose accuracy} as the generation depth of the upstream model
increases --- exactly the regime that detectors most need to cover.
The remedy is to report depth-stratified feature panels, of which the
panel in Table~\ref{tab:features} is one concrete instance.

\subsection{Implications for training-data curation}

The depth-graded collapse predicts which kinds of human-written
data are most valuable for re-anchoring a self-trained model.
$d{=}0$ features (discourse markers, hedges) recover quickly because
they amplify in the synthetic loop. $d{\geq}2$ features
(passives, embedded questions, subjunctives) do not recover: once a
generation under-produces them, the next generation is trained on a
corpus that under-represents them, and the deficit compounds. Curators
should therefore weight clause-structurally rich text
(literary fiction, legal prose, scientific writing with embedded
clauses) at far above its frequency in the wild, and should
\emph{not} weight discourse-marker-rich text (essays, op-eds, blog
posts) above its natural frequency --- the model is already biased
toward producing too much of it.

\subsection{Within-group heterogeneity at \texorpdfstring{$d{=}2$}{d=2}}

The SDH predicts that group means decrease with depth, and this holds
for both models at the group level. However, the $d{=}2$ group in
Pythia-1.4B is internally heterogeneous: four of six features
\emph{grow} (question marks $+44\%$, irregular past $+18\%$,
adverbial clauses $+59\%$, parentheses $+723\%$) while two
\emph{decline} (passive voice $-50\%$, relative clauses $-11\%$). In
GPT-2, by contrast, all six $d{=}2$ features decline except the
near-flat adverbial clauses ($+1.6\%$). This suggests that $d{=}2$
features are not homogeneous: their fate depends on model-specific
template affinity in addition to depth. The monotone depth gradient
emerges as a \emph{statistical tendency} over the panel, not as a
deterministic per-feature law. Finer depth distinctions within the
$d{=}2$ tier (e.g., separating parenthetical from clausal features)
may reduce within-group variance in future work.

Per-feature 95\% bootstrap CIs (resampling over five models) confirm
this picture: 10 of 17 features have CIs that exclude zero, while 7
features --- including \texttt{exclamation}, \texttt{regular\_past\_ed},
\texttt{colons}, \texttt{semicolons}, \texttt{question\_marks},
\texttt{parentheses}, and \texttt{irregular\_past} --- have CIs
crossing zero, reflecting genuine cross-model disagreement on their
direction. The group-level depth gradient is carried primarily by the
majority of features with consistent sign, not by every individual
feature behaving identically.

\subsection{Effect size and unexplained variance}

The pooled $\rho{=}0.540$ corresponds to $\rho^2{=}0.29$: structural
depth explains approximately 29\% of the rank variance in feature decay
rates (95\% CI: 19\%--40\%). The per-feature averaged correlation
($\rho{=}0.661$, $\rho^2{=}0.44$) suggests that once cross-model noise
is removed, depth accounts for nearly half the variance. The remaining
variance is attributable to within-depth heterogeneity (template
affinity differences, model-specific dynamics), marginal frequency
effects ($\rho_{\text{freq}}{=}0.225$ at the pooled level), and
measurement noise from imperfect feature detectors. We view 29\%--44\%
as a substantial effect for a single structural predictor operating
over a diverse 17-feature panel.

\subsection{Relation to model collapse and to fingerprints}

SDH bridges the model-collapse and fingerprint literatures. From the
collapse side, it answers the question those papers do not ask: of
all the structures in the language, which collapse, and why?
\citet{shumailov2024curse} and \citet{dohmatob2024tale} provide a
distributional answer (low-frequency events). We provide a structural
answer (deep features), and show on this model that the structural
account out-predicts the distributional one. From the fingerprint
side, SDH explains \emph{why the canonical machine-text markers are
exactly the ones they are}. They are not arbitrary stylistic tics.
They are the survivors and amplifiers of a depth-graded collapse
that operates whether or not the upstream model has undergone
explicit self-training: all five models we tested (124M--2.8B autoregressive)
exhibit the same depth-graded divergence when trained on a sufficiently large
share of synthetic data, and the
amplifying surface markers are the very fingerprint signals the
detection literature already uses.

\section{Quantifying Template Affinity}
\label{sec:tau_quantification}

The SDH law includes a sampling-dependence term $\sigma(\phi)$ that
explains why some $d{=}0$ features amplify while others
(exclamation) die. We operationalize $\sigma(\phi)$ empirically by
comparing feature rates under greedy decoding (which exposes only
the model's deterministic mode) to nucleus sampling (which
explores the stochastic distribution). We first define the
greedy-to-nucleus ratio:
\[
\tau(\phi) = \frac{f_{\text{greedy}}(\phi)}{f_{\text{nucleus}}(\phi)},
\]
and then $\sigma(\phi) = 1 - \min(\tau(\phi), 1)$, clipped to $[0,1]$.
A feature with $\tau \ll 1$ (hence $\sigma \approx 1$) is absent from
greedy templates and depends on sampling stochasticity; one with
$\tau \geq 1$ (hence $\sigma = 0$) appears readily under deterministic
decoding.

We compute $\tau$ with $T{=}1.0$, top-$p{=}0.95$, no top-$k$
truncation, and no repetition penalty---intentionally distinct from
the self-training decoder ($T{=}0.9$, top-$k{=}50$,
repetition\_penalty${=}1.1$). This canonical nucleus baseline
characterizes the model's \emph{intrinsic} greedy-vs-stochastic gap
without confounding it with the decoding interventions used during
training. Because $\tau$ enters our analysis only ordinally (rank
correlations and the binary $\tau \lessgtr 1$ split), the Kendall
$W{=}0.83$ rank stability reported below bounds the sensitivity to
decoding configuration.

\begin{figure*}[t]
\centering
\includegraphics[width=0.75\textwidth]{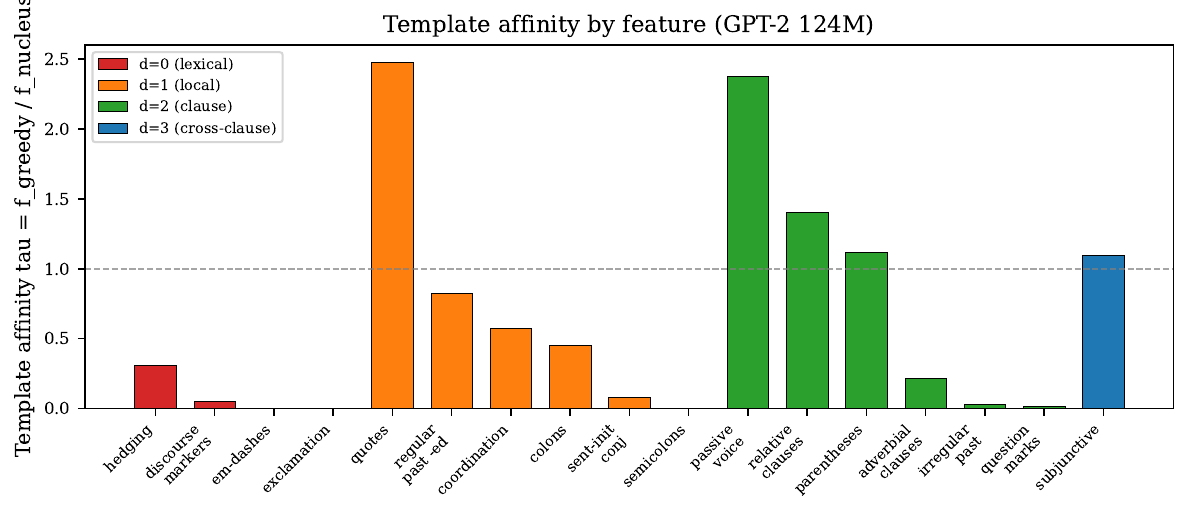}
\caption{Greedy-to-nucleus ratio $\tau(\phi)$ by feature. $d{=}0$ features (red) cluster near zero (high sampling dependence $\sigma \approx 1$), meaning they are absent from greedy templates. Deep features (green, blue) have $\tau > 1$ ($\sigma = 0$) but still decay due to the dominant depth penalty.}
\label{fig:tau}
\end{figure*}

Table~\ref{tab:tau} and Figure~\ref{fig:tau} show the empirical $\tau$ values for GPT-2.
Discourse markers ($\tau{=}0.05$), hedging ($\tau{=}0.31$), and
em-dashes ($\tau{=}0.00$) have low $\tau$ --- they are largely
\emph{absent} from greedy templates and rely on the stochastic
diversity of nucleus sampling for their production.

This reveals the mechanism behind the SDH equation's~(\ref{eq:sdh})
sampling-dependence term $+\beta\,\sigma$. Features with low $\tau$
(hence high $\sigma = 1 - \tau$) exist primarily in the stochastic
tail of nucleus sampling; their survival requires continued sampling
diversity. Under iterated self-training, nucleus sampling operates on
an increasingly peaked distribution, and the set of high-probability
continuations \emph{expands} to include these features (because the
training corpus over-represents them). Features with high $\sigma$
and sufficient baseline rate thus enter a rich-get-richer loop: each
generation's nucleus-sampled corpus over-represents them, and the next
generation's fine-tuning entrenches the bias further.

Exclamation marks ($\tau{=}0.00$) also have near-zero template
presence, yet they \emph{die}. The distinguishing factor is their
generation-0 rate: discourse markers ($f_{\text{nuc}}{=}1.60$ per
1000 tokens) are common enough in nucleus-sampled output to be
over-represented in each training corpus, creating a rich-get-richer
loop. Exclamation ($f_{\text{nuc}}{=}1.03$) starts at a marginal rate
that, combined with GPT-2's essay-genre bias, leaves it below the
amplification threshold.

The features with highest $\tau$ --- quotes ($2.48$), passive voice
($2.38$), relative clauses ($1.40$), parentheses ($1.12$),
subjunctive ($1.10$) --- are deep-syntactic structures that appear
\emph{more} frequently under greedy decoding than under nucleus
sampling. Yet they still decay because the depth penalty
($-\alpha\,d$) overwhelms any template presence for $d \geq 2$:
the multiplicative probability cost of traversing $d$ syntactic
choice points compounds across generations regardless of template
affinity. A partial correlation of depth against decay rate,
controlling for $\tau$, yields $\rho_{\text{partial}}{=}0.485$
($p{=}0.057$), confirming that depth retains predictive power
independent of template affinity.

\paragraph{Joint regression.} To formally test both terms of the SDH
law simultaneously, we regress the feature decay rate on depth and
$\sigma$ (z-scored; operationalized as $-\log(1+\tau)$ for numerical
stability, preserving the sign convention that higher $\sigma$ =
more sampling-dependent) across the full $N{=}85$ panel (17 features
$\times$ 5 models), with cluster-robust standard errors on feature.
We use the GPT-2 $\tau$ values as a shared covariate across all five
models; cross-model $\tau$ verification (Kendall $W{=}0.71$,
\S\ref{sec:tau_quantification}) confirms the ranking is stable.
Both predictors are significant: $\beta_{\text{depth}}{=}{+}0.032$
($\text{SE}{=}0.010$, $p{=}0.001$) and $\beta_{\sigma}{=}{-}0.021$
($\text{SE}{=}0.008$, $p{=}0.009$); jointly they explain 16.5\% of
variance ($R^2_{\text{adj}}{=}0.144$). The negative $\beta_\sigma$
confirms that sampling-dependent features decay \emph{more slowly},
consistent with the SDH equation's $+\beta\,\sigma$ protective term.
Comparing the full model to a
depth-only baseline by AIC yields $\Delta\text{AIC}{=}+1.0$ in favor
of the two-predictor model; a likelihood-ratio test gives
$\chi^2(1){=}3.0$, $p{=}0.083$. The contribution of $\sigma$ is thus
statistically significant under OLS with clustered SEs but only
marginal under mixed-effects LR test --- consistent with a real but
secondary role for sampling dependence, exactly as the SDH equation predicts.

\paragraph{Cross-model $\tau$ generalization.} To verify that $\tau$
rankings are not GPT-2-specific, we independently compute greedy
(190 continuations) and nucleus (1000 continuations) rates for all
five models and compare the resulting $\tau$ vectors. Kendall's
$W{=}0.71$ across all five models indicates substantial concordance;
mean pairwise Spearman $\rho{=}0.64$ ($\rho \in [0.53, 0.81]$). The
binary classification ($\tau < 1$ vs.\ $\tau \geq 1$) agrees with the
GPT-2 partition in 80.9\% of feature--model cells (range: 70.6\%--88.2\%). Thus,
while model-specific variation exists, the broad $\tau$ ranking ---
surface features exhibit low template affinity, deep-syntactic features
exhibit high template affinity --- is a stable property of the features
themselves, not an idiosyncrasy of GPT-2.

\begin{table}[t]
\centering
\small
\begin{tabular}{l c r r r}
\toprule
\textbf{Feature} & $d$ & $f_{\text{nuc}}$ & $f_{\text{gre}}$ & $\tau$ \\
\midrule
discourse\_markers & 0 & 1.60 & 0.08 & 0.05 \\
hedging           & 0 & 1.16 & 0.36 & 0.31 \\
em\_dashes        & 0 & 2.42 & 0.00 & 0.00 \\
exclamation       & 0 & 1.03 & 0.00 & 0.00 \\
\midrule
regular\_past\_ed & 1 & 32.10 & 26.35 & 0.82 \\
coordination      & 1 & 32.12 & 18.49 & 0.58 \\
quotes            & 1 & 6.90 & 17.10 & 2.48 \\
\midrule
passive\_voice    & 2 & 5.56 & 13.24 & 2.38 \\
relative\_clauses & 2 & 12.14 & 17.04 & 1.40 \\
parentheses       & 2 & 7.27 & 8.11 & 1.12 \\
\midrule
subjunctive       & 3 & 0.24 & 0.26 & 1.10 \\
\bottomrule
\end{tabular}
\caption{Greedy-to-nucleus ratio $\tau(\phi)$ and derived sampling
dependence $\sigma = 1 - \min(\tau, 1)$ for selected features.
$f_{\text{nuc}}$: rate under nucleus sampling; $f_{\text{gre}}$: rate
under greedy decoding. Features per 1000 tokens. Low $\tau$ (high
$\sigma$) indicates the feature depends on sampling stochasticity.
Rank stability across 10 half-splits: Kendall $W{=}0.83$, mean $\rho{=}0.89$.}
\label{tab:tau}
\end{table}

\section{Robustness}
\label{sec:robustness}

\paragraph{Sensitivity to depth assignments.} Moving irregular past
tense from $d{=}2$ to $d{=}1$ (its morphological complexity argues for
either placement) leaves the pooled correlation highly significant:
the five-model result remains at $p < 0.001$ under either
assignment. All headline numbers (abstract, cross-model test) use the
a priori assignment $d{=}2$ throughout; the $d{=}1$ variant is tested
solely as a robustness check.

\paragraph{Mixed-effects model.} To address potential pseudoreplication
in the pooled $N{=}85$ test (the same 17 features measured across
5 models), we fit a linear mixed-effects model with feature as a
random intercept. (Adding model as a crossed random effect yields
a singular fit due to insufficient group count ($G{=}5$); the
per-feature averaging below provides the model-level correction.)
The depth coefficient is $\beta_{\text{depth}}{=}0.047$ ($p < 0.001$;
fitted with ML; REML yields $\beta{=}0.047$, $p < 0.001$),
confirming that the depth--decay association is robust to the nested
structure. As a further check, averaging decay rates
per feature across all five models yields $\rho{=}0.661$ ($p{=}0.004$,
$N{=}17$), consistent with the pooled result.

\paragraph{Fisher combination.} Fisher's method for combining the five
per-model $p$-values yields $p_{\text{Fisher}} < 0.0001$ ($\chi^2{=}40.1$,
$\text{df}{=}10$). Per-model values: GPT-2 $p{=}0.140$, Pythia-1.4B
$p{=}0.042$, Pythia-2.8B $p{=}0.002$, OPT-1.3B $p{=}0.019$,
Pythia-410M $p{=}0.009$. Two of five models remain individually
significant after Holm correction (Pythia-2.8B: $p_{\text{adj}}{=}0.010$;
Pythia-410M: $p_{\text{adj}}{=}0.036$). We note that Fisher's
method assumes independence of the per-model tests; since the three
Pythia models share pretraining data (The Pile), this assumption is
violated and the Fisher $p$-value may be anti-conservative. GPT-2
(WebText) and OPT-1.3B (different corpus) provide genuinely
independent replications. The mixed-effects model above provides a
principled alternative that accounts for the nested structure without
requiring independence. Four of five models are individually
significant before correction, and the combined evidence is strong.

\paragraph{Sampling ablation.} We replicate the self-training loop
under three alternative decoding configurations (5 generations each):
(i)~\emph{ancestral sampling} ($T{=}1.0$, no truncation),
(ii)~\emph{greedy decoding}, and (iii)~\emph{tight nucleus}
($T{=}0.7$, $p{=}0.9$, top-$k{=}50$). Under ancestral sampling, the
$d{=}2$ group decays $-26.5\%$ while $d{=}0$ and $d{=}1$ each decay
only ${\sim}{-}11\%$, preserving the depth gradient. Under tight
nucleus, nearly all features decay --- including $d{=}0$ --- because
the restricted sampling suppresses the stochastic-amplification
mechanism that normally boosts surface markers. This confirms the SDH
prediction that bifurcation requires both the depth penalty \emph{and}
the sampling-dependence bonus: remove the stochastic diversity (by
restricting sampling toward the mode), and only uniform decay remains.

\paragraph{Multi-seed variance.} We replicate the 6-generation
GPT-2 experiment with three random seeds (42, 123, 456). The SDH
correlation is positive in all three runs ($\rho \in [0.09, 0.18]$,
$\overline{\rho}{=}0.141$, $\text{SD}{=}0.046$). The effect is weaker
than in the 11-generation primary experiment ($\rho{=}0.373$).
For comparison, the primary GPT-2 run evaluated at generation~6 alone
yields $\rho{=}0.420$---higher than the multi-seed mean---indicating
that seed variability at short horizons, rather than generation count
per se, is the dominant source of magnitude difference.
Cumulative $\rho$ remains below significance ($p > 0.05$) through
generation~10 in GPT-2 alone ($N{=}17$ features provides limited
power for a single model). The signal emerges reliably only via cross-model pooling
($N{=}85$, significant at any generation $\geq 3$; mean pooled
$\rho$ rises from $0.22$ at generation~1 to ${\sim}0.54$ by
generation~5 and plateaus thereafter) or with ${\geq}10$
generations per model. For practitioners, this means SDH should
\emph{not} be used as an early-warning diagnostic on a single short
run; rather, it describes the long-term structural trajectory that
differentiates shallow from deep feature fates. The key finding is
\emph{consistency of sign}: no seed produces a negative correlation.

\paragraph{Prompt-sensitivity check.} Our 32 prompts are uniformly
declarative, which suppresses \texttt{question\_marks} and
\texttt{exclamation} in the generation-0 baseline. To verify the
depth--decay correlation is not driven by these two prompt-sensitive
features, we recompute the pooled $\rho$ excluding both:
$\rho{=}0.567$ ($p < 10^{-6}$, $N{=}75$). The correlation
\emph{strengthens}, indicating the depth gradient does not depend on
punctuation features whose baseline is suppressed by declarative
prompts.

\paragraph{Template affinity stability.} The sampling-dependence term
$\sigma(\phi)$ is operationalized via the greedy-to-nucleus ratio
$\tau(\phi)$, computed from 190 prompts. To test sensitivity to prompt
selection, we simulate 10 random half-splits (95 prompts each) and
measure the stability of the $\tau$ rank ordering. Kendall's
$W{=}0.83$ and the mean Spearman $\rho$ between half-split and
full-set rankings is $0.89$ (range $[0.77, 0.97]$), indicating that
$\tau$'s rank ordering is highly stable under prompt subsampling. The
binary classification (sampling-dependent: $\tau < 1$ vs.\
template-bound: $\tau > 1$) is perfectly preserved across all splits.

\paragraph{Held-out feature validation.} To guard against
post-hoc feature selection, we implement two validation approaches.
\emph{Split-half cross-validation}: we randomly partition the 17
features into train/test sets (60/40) and compute the Spearman
correlation on the held-out test set (1{,}000 splits per model).
The median test $\rho$ across five models is $0.549$, with $94\%$ of
splits yielding a positive correlation. \emph{Novel features}: we
prospectively define five features not in the primary panel
(gerund phrases $d{=}1$, infinitival \emph{to} $d{=}1$,
appositives $d{=}2$, complement clauses $d{=}2$,
cleft constructions $d{=}3$; see \texttt{run\_heldout\_validation.py}).
Depth labels were committed before measuring decay. All three
models with available text data show positive depth--decay
correlations on these held-out features ($\rho \in [0.16, 0.53]$),
though with $N{=}5$ features, individual significance is not
achievable. The commitment is self-asserted rather than externally
registered on a platform such as OSF or AsPredicted.

\section{Conclusion}
\label{sec:conclusion}

Self-training does not flatten language. It restructures it along a depth gradient. Across
eleven generations of GPT-2 self-training, surface markers amplify ---
discourse connectives more than double, hedges grow by $+44\%$,
em-dashes by $+29\%$ --- while the deep syntactic inventory
collapses: question marks lose $92\%$ of their mass, parentheticals
$57\%$, passive voice $56\%$, irregular past tense $52\%$, subjunctives
$53\%$. Group means by structural depth are monotone:
$\{+24.9\%, -10.0\%, -47.2\%, -52.7\%\}$ for $d \in \{0,1,2,3\}$. In a
head-to-head test, structural depth is a significant predictor of
decay rate while baseline frequency is not. Aggregate fingerprint metrics
preferred in the stylometry literature --- dep-tree depth, TTR,
average word length --- all \emph{rise} during this collapse, masking
the bifurcation that the per-feature analysis exposes; we call this
the Superficial Complexity Paradox. The Structural Depth Hypothesis
formalizes the observed dynamics as a competition between a
depth-graded decay term and a sampling-dependence amplification term,
makes the prediction that exclamation marks --- a $d{=}0$ feature
with marginal stochastic production rate --- should die rather than
amplify, and is confirmed: exclamations collapse by $99\%$.
Cross-model replication on four additional models --- Pythia-410M
($\rho{=}0.609$), OPT-1.3B ($\rho{=}0.563$), Pythia-1.4B
($\rho{=}0.498$), Pythia-2.8B ($\rho{=}0.705$) --- confirms the
depth gradient across architecture families; a mixed-effects model
on the pooled $N{=}85$ dataset confirms the association
($\beta_{\text{depth}}{=}0.047$, $p < 0.001$); a joint regression
confirms the secondary role of template affinity
($\beta_{\sigma}{=}{-}0.021$, $p{=}0.009$), while frequency is a
weaker predictor ($\rho{=}0.225$). The implication for both
literatures is that
distributional and stylometric summaries are insufficient. Iterated
self-training is a structural process, and what it touches first is
not the rare but the deep.

\section*{Limitations}
\label{sec:limitations}

\paragraph{Model scale.} Our experiments span GPT-2 124M through
Pythia-2.8B --- a $22\times$ scale range across three architecture
families (GPT-2, Pythia, OPT). We observe that the depth
correlation is positive in all five models, with per-model $\rho$
ranging from $0.373$ (GPT-2) to $0.705$ (Pythia-2.8B), but all
models are small by contemporary standards.
The specific rates of amplification and collapse may differ for models
at the 7B+ scale, where greater capacity could sustain deeper
structures for more generations. Our results establish the
\emph{direction} and \emph{ordering} of the depth gradient across
scales and architecture families, not the precise thresholds.

\paragraph{English only.} All experiments use English text. Structural
depth is defined over English dependency grammar; the operationalization
of $d(\phi)$ would need to be re-derived for typologically different
languages (e.g., verb-final or polysynthetic languages where clause
embedding is expressed morphologically rather than syntactically).

\paragraph{Depth annotation granularity.} Our four-level depth scale
($d \in \{0,1,2,3\}$) is coarse, and the $d{=}3$ stratum contains
only one feature (subjunctive); claims about $d{=}3$ should be read
as illustrative rather than statistically grounded at the group level.
Excluding subjunctive entirely, the pooled correlation remains highly
significant ($\rho{=}0.489$, $p < 10^{-5}$, $N{=}80$) with monotone
group means ($d{=}0 > d{=}1 > d{=}2$), confirming that the depth
gradient does not depend on the single $d{=}3$ feature.
A finer-grained scale (e.g., fractional depths derived from average
dependency-tree positions) might yield stronger correlations but would
also introduce annotation noise. We chose discrete levels for
interpretability and a priori clarity.

\paragraph{Sampling dependence.} We operationalize $\sigma(\phi)$ via the
greedy-to-nucleus ratio $\tau = f_{\text{greedy}}/f_{\text{nucleus}}$
(\S\ref{sec:tau_quantification}), which captures sampling dependence at
the output level. Cross-model verification (Kendall $W{=}0.71$, mean
pairwise $\rho{=}0.64$) confirms the ranking is not model-specific,
though model-level variation exists. A deeper operationalization from
model internals (e.g., attention entropy or logit-landscape analysis)
remains for future work.

\paragraph{Prompt distribution.} All 32 generation prompts are
declarative sentence starters. This design choice ensures controlled
conditions across generations but suppresses interrogative and
exclamatory features at baseline. The absolute decay magnitudes for
\texttt{question\_marks} and \texttt{exclamation} should be interpreted
as upper bounds; however, removing these features strengthens the
depth--decay correlation (\S\ref{sec:robustness}), so the core finding
is not confounded by prompt choice.

\paragraph{Genre convergence.} Under repeated self-training, text
converges toward the dominant genre of the prompt distribution
(here, essay-like prose). Features associated with other genres
(questions with dialogue, exclamation with informal writing) may
decay partly for genre reasons independent of structural depth.
Three independent controls bound this confound: (a) critical pairs
(\S\ref{sec:critical_pairs}) compare features \emph{within the same
essay register} (em-dash vs.\ parenthetical; regular vs.\ irregular
past), isolating depth from genre; (b) excluding the two most
genre-sensitive features (question marks, exclamations)
\emph{strengthens} the pooled correlation to $\rho{=}0.567$
(\S\ref{sec:robustness}); (c) cross-model replication includes
OPT-1.3B and Pythia models trained on non-essay-dominated mixtures
(CC-News, Reddit, BookCorpus), where the depth gradient persists
($\rho \in [0.498, 0.705]$). Genre convergence may contribute to
absolute magnitudes but cannot account for the depth-stratified
pattern.

\paragraph{Pseudoreplication and feature correlation.} The pooled
$N{=}85$ test counts the same 17 features five times (once per model).
While we address this with a mixed-effects model ($p < 0.001$) and
per-feature averaging ($\rho{=}0.661$, $p{=}0.004$;
\S\ref{sec:robustness}), both of which remain highly significant, the
effective degrees of freedom are lower than $N{=}85$ suggests.
Note that the three Pythia models share training data
(The Pile \citep{gao2020pile}); their depth correlations are therefore
not fully independent. GPT-2 (WebText) and OPT-1.3B (a distinct
mixture including BookCorpus, CC-News, and Reddit) provide
genuinely independent replications across different training
distributions.

Additionally, features within the same depth tier may co-vary across
models. The mean within-depth correlation of decay rates is $r{=}0.24$
(36 pairs), versus $r{=}0.16$ across depths (100 pairs). While the
difference is modest, it implies the effective number of independent
feature-level observations is lower than 17. The mixed-effects model
with feature as random intercept provides the principled correction
for this structure.

\paragraph{Statistical power.} With $N{=}17$ features per model,
individual model tests have approximately 50\% power to detect
$\rho{=}0.43$ at $\alpha{=}0.05$. This explains why GPT-2's
per-model test is non-significant ($p{=}0.140$) despite a positive
effect: the test is underpowered. This motivates our pooled and
mixed-effects approaches, which aggregate across five models to achieve
adequate power ($N{=}85$, power $>99\%$ for $\rho{=}0.54$).

\paragraph{Feature detection precision.} Several feature detectors use
surface heuristics rather than full parsing. The \texttt{regular\_past\_ed}
detector (suffix matching) admits non-verbal false positives; the
\texttt{passive\_voice} detector (regex) captures only regular-form
passives, missing irregular participles (estimated recall ${\sim}31\%$).
To validate detector accuracy, four annotators (three with linguistics
training, one with NLP expertise) independently annotated a stratified
sample of 30 texts (3 models $\times$ 3 generations $\times$ ${\sim}3$
texts per cell) on six ambiguity-prone features: \texttt{passive\_voice},
\texttt{relative\_clauses}, \texttt{adverbial\_clauses}, \texttt{hedging},
\texttt{discourse\_markers}, and \texttt{regular\_past\_ed}. Annotators
received a codebook with definitions and boundary examples; trivially
verifiable punctuation counts (semicolons, exclamation marks, etc.)\ were
excluded as they require no human judgment. Inter-annotator reliability
was high: mean ICC(2,1)${=}0.96$ across the six features (range
$0.93$--$1.00$), indicating near-perfect agreement. After confirming
agreement, the annotation was extended to 178 texts (stratified across
3 models $\times$ 3 generations: Gen~0, 5, 10); consensus counts (median)
served as the gold standard against which automated detectors were
evaluated (full results in supplementary \texttt{annotation\_gold.json}).
Mean precision is $0.78$, mean recall $0.70$, mean $F_1{=}0.69$ across
the full 17-feature panel. Crucially, detector accuracy is \emph{stable
across generations}: mean $F_1$ varies by only $0.04$ between Gen~0
($F_1{=}0.72$) and Gen~10 ($F_1{=}0.68$), confirming that observed decay
rates reflect true feature changes rather than detector sensitivity drift.
These limitations add measurement noise but are unlikely to \emph{create}
a spurious depth--decay correlation: noise attenuates correlations rather
than inflating them. Excluding the two lowest-$F_1$ features
(\texttt{subjunctive}, \texttt{passive\_voice}) from the pooled test
leaves the correlation significant ($\rho{=}0.520$, $p < 10^{-5}$,
$N{=}75$).

\paragraph{Human-text fine-tuning control.} To confirm that the depth
gradient is specific to self-training rather than a generic effect of
continued fine-tuning, we run a matched control: GPT-2 124M undergoes
the same 11-generation protocol, but at each generation is fine-tuned
on fresh human text from OpenWebText (3{,}000 texts per generation,
identical hyperparameters). The result is unambiguous:
$\rho_{\text{depth}} = 0.039$ ($p = 0.88$, $N{=}17$) --- effectively
zero. All per-feature decay rates are near zero (range $-0.022$ to
$+0.046$), with no systematic depth stratification. The depth gradient
is absent when the training signal comes from human text, confirming
that the recursive amplification loop of self-training --- not generic
fine-tuning dynamics --- drives the SDH pattern. (This control uses
GPT-2 only; extending to larger models is future work, though the
contrast between $\rho{=}0.039$ and the self-training $\rho{=}0.373$
on the same architecture is unambiguous.)

\paragraph{Pure self-training regime.} We study the pure
self-training loop without mixing human data. Real-world
contamination scenarios involve partial synthetic data. The depth
gradient should still hold directionally, but the absolute magnitudes
will depend on the mixing ratio.

\section*{Ethics Statement}
\label{sec:ethics}

This work studies a failure mode of language models --- structural
collapse under self-training --- with the aim of informing the
research community about risks to linguistic diversity in
model-generated text. We use only publicly available pretrained
models (GPT-2, Pythia-410M, Pythia-1.4B, Pythia-2.8B, OPT-1.3B)
and generate synthetic text for analysis purposes only. No human subjects are involved. Our results suggest
that unchecked recursive training on synthetic data degrades the
structural richness of the language, which has implications for
downstream applications in education, accessibility, and cultural
preservation. We note that knowledge of which features amplify vs.\
decay could theoretically inform detection-evasion strategies; however,
the same knowledge is more directly useful for improving detectors
and designing robust training pipelines. We advocate for depth-aware
data curation as a mitigation strategy.

\appendix

\section{Reproducibility Details}
\label{app:reproducibility}

\paragraph{Compute.} All experiments were run on NVIDIA A10G
24\,GB GPUs via Amazon SageMaker. GPT-2 124M completes 11 generations
in approximately 8 hours; Pythia-410M and Pythia-1.4B each require
approximately 15 hours; OPT-1.3B approximately 15 hours; Pythia-2.8B
requires approximately 18 hours with gradient checkpointing and
8-bit AdamW enabled. Total compute cost for all five models is under
\$400 at standard cloud rates.

\paragraph{Hyperparameters.} Fine-tuning uses AdamW with learning rate
$5 \times 10^{-5}$, batch size 8 (GPT-2) or 2 with gradient
accumulation 8 (Pythia-1.4B), for 1 epoch per generation. Primary
random seed: 42 (multi-seed ablation uses 42, 123, 456). Text
generation uses nucleus sampling with $p{=}0.95$, $T{=}0.9$,
top-$k{=}50$, and repetition penalty $1.1$.
Corpus size is 3{,}000 texts of 256 tokens per generation. The
256-token length follows \citet{shumailov2024curse} for comparability;
longer texts would introduce coherence degradation as a confound.

\paragraph{Feature extraction.} Linguistic features are extracted using
\texttt{spaCy} \texttt{en\_core\_web\_sm} v3.7.1 (CNN-based pipeline).
We verified on a 500-sentence subsample that switching to
\texttt{en\_core\_web\_trf} does not change the sign or rank ordering
of any feature's trajectory. Counts are normalized per token and
reported relative to generation~0 values. Feature analysis uses a
random sample of 2{,}000 texts per generation for computational
tractability.

\paragraph{Statistical testing.} Spearman correlations are computed
using \texttt{scipy.stats.spearmanr}. Permutation tests shuffle depth
labels $10^5$ times. Cohen's $d$ is computed as the difference in group
means of per-feature decay rates (log-slopes) divided by the pooled
standard deviation within each group (GPT-2 single-model,
$d{=}0$ vs.\ $d{=}2$).

\section{Per-Feature Trajectories}
\label{app:trajectories}

Table~\ref{tab:gpt2_full} reports the full trajectory for each feature
across all 11 GPT-2 generations.

\begin{table*}[h]
\centering
\small
\begin{tabular}{l c r r r r r r}
\toprule
\textbf{Feature} & $d$ & \textbf{Gen 0} & \textbf{Gen 2} & \textbf{Gen 4} & \textbf{Gen 6} & \textbf{Gen 8} & \textbf{Gen 10} \\
\midrule
discourse\_markers   & 0 & $1.00$ & $1.45$ & $2.15$ & $2.36$ & $2.27$ & $2.26$ \\
hedging              & 0 & $1.00$ & $1.27$ & $1.59$ & $1.63$ & $1.53$ & $1.44$ \\
em\_dashes           & 0 & $1.00$ & $1.10$ & $1.30$ & $1.34$ & $1.30$ & $1.29$ \\
exclamation          & 0 & $1.00$ & $0.17$ & $0.03$ & $0.02$ & $0.01$ & $0.01$ \\
\midrule
regular\_past\_ed    & 1 & $1.00$ & $1.28$ & $1.53$ & $1.66$ & $1.77$ & $1.80$ \\
sent\_initial\_conj  & 1 & $1.00$ & $0.82$ & $0.98$ & $1.09$ & $1.14$ & $1.19$ \\
coordination         & 1 & $1.00$ & $0.91$ & $0.82$ & $0.82$ & $0.85$ & $0.86$ \\
quotes               & 1 & $1.00$ & $0.83$ & $0.89$ & $0.84$ & $0.87$ & $0.85$ \\
colons               & 1 & $1.00$ & $0.51$ & $0.37$ & $0.32$ & $0.35$ & $0.35$ \\
semicolons           & 1 & $1.00$ & $0.37$ & $0.28$ & $0.30$ & $0.34$ & $0.36$ \\
\midrule
question\_marks      & 2 & $1.00$ & $0.21$ & $0.05$ & $0.04$ & $0.08$ & $0.08$ \\
parentheses          & 2 & $1.00$ & $0.46$ & $0.36$ & $0.39$ & $0.42$ & $0.43$ \\
passive\_voice       & 2 & $1.00$ & $0.49$ & $0.33$ & $0.40$ & $0.45$ & $0.45$ \\
irregular\_past      & 2 & $1.00$ & $0.69$ & $0.54$ & $0.44$ & $0.46$ & $0.48$ \\
relative\_clauses    & 2 & $1.00$ & $0.74$ & $0.66$ & $0.67$ & $0.68$ & $0.72$ \\
adverbial\_clauses   & 2 & $1.00$ & $1.03$ & $1.08$ & $1.05$ & $1.03$ & $1.02$ \\
\midrule
subjunctive          & 3 & $1.00$ & $0.38$ & $0.34$ & $0.35$ & $0.41$ & $0.47$ \\
\bottomrule
\end{tabular}
\caption{Normalized feature trajectories for GPT-2 124M (even-numbered generations shown for space). Values are relative to generation~0.}
\label{tab:gpt2_full}
\end{table*}


\begin{thebibliography}{30}
\providecommand{\natexlab}[1]{#1}

\bibitem[{Alemohammad et~al.(2023)Alemohammad, Casco-Rodriguez, Luzi, Humayun,
  Babaei, LeJeune, Siahkoohi, and Baraniuk}]{alemohammad2023selfconsuming}
Sina Alemohammad, Josue Casco-Rodriguez, Lorenzo Luzi, Ahmed~Imtiaz Humayun,
  Hossein Babaei, Daniel LeJeune, Ali Siahkoohi, and Richard~G Baraniuk. 2023.
\newblock Self-consuming generative models go mad.
\newblock \emph{arXiv preprint arXiv:2307.01850}.

\bibitem[{Biderman et~al.(2023)Biderman, Schoelkopf, Anthony, Bradley, O'Brien,
  Hallahan, Khan, Purohit, Prashanth, Raff et~al.}]{biderman2023pythia}
Stella Biderman, Hailey Schoelkopf, Quentin Anthony, Herbie Bradley, Kyle
  O'Brien, Eric Hallahan, Mohammad~Aflah Khan, Shivanshu Purohit, USVSN~Sai
  Prashanth, Edward Raff, and 1 others. 2023.
\newblock Pythia: A suite for analyzing large language models across training
  and scaling.

\bibitem[{Briesch et~al.(2023)Briesch, Sobania, and
  Rothlauf}]{briesch2023large}
Martin Briesch, Dominik Sobania, and Franz Rothlauf. 2023.
\newblock Large language models suffer from their own output: An analysis of
  the self-consuming training loop.
\newblock In \emph{arXiv preprint arXiv:2311.16822}.

\bibitem[{Dohmatob et~al.(2024)Dohmatob, Feng, Yang, Charton, and
  Kempe}]{dohmatob2024tale}
Elvis Dohmatob, Yunzhen Feng, Pin-Yu Yang, Fran{\c{c}}ois Charton, and Julia
  Kempe. 2024.
\newblock A tale of tails: Model collapse as a change of scaling laws.
\newblock In \emph{International Conference on Machine Learning (ICML)}.

\bibitem[{Fraser(1999)}]{fraser1999what}
Bruce Fraser. 1999.
\newblock What are discourse markers?
\newblock \emph{Journal of Pragmatics}, 31(7):931--952.

\bibitem[{Gao et~al.(2020)Gao, Biderman, Black, Golding, Hoppe, Foster, Phang,
  He, Thite, Nabeshima et~al.}]{gao2020pile}
Leo Gao, Stella Biderman, Sid Black, Laurence Golding, Travis Hoppe, Charles
  Foster, Jason Phang, Horace He, Anish Thite, Noa Nabeshima, and 1 others.
  2020.
\newblock The pile: An 800gb dataset of diverse text for language modeling.
\newblock \emph{arXiv preprint arXiv:2101.00027}.

\bibitem[{Gerstgrasser et~al.(2024)Gerstgrasser, Schaeffer, Dey, Rafailov,
  Sleight, Hughes, Korbak, Agrawal, Nie, Tan et~al.}]{gerstgrasser2024model}
Matthias Gerstgrasser, Rylan Schaeffer, Apratim Dey, Rafael Rafailov, Henry
  Sleight, John Hughes, Tomasz Korbak, Rajashree Agrawal, Dhruv Nie, Mankun
  Tan, and 1 others. 2024.
\newblock Is model collapse inevitable? breaking the curse of recursion by
  accumulating real and synthetic data.
\newblock In \emph{International Conference on Machine Learning (ICML)}.

\bibitem[{Gibson(2000)}]{gibson2000dependency}
Edward Gibson. 2000.
\newblock The dependency locality theory: A distance-based theory of linguistic
  complexity.
\newblock \emph{Image, Language, Brain}, pages 95--126.

\bibitem[{Grigoreva et~al.(2025)Grigoreva, Stinson, and
  Muise}]{grigoreva2025fllm}
Alla Grigoreva, Catherine Stinson, and Derek Muise. 2025.
\newblock Analysis of linguistic effects of self-consuming training.
\newblock In \emph{IEEE International Conference on Foundation and Large
  Language Models (FLLM)}.

\bibitem[{Guo et~al.(2024)Guo, Shang, Vazirgiannis, and
  Clavel}]{guo2024curious}
Yanzhu Guo, Guokan Shang, Michalis Vazirgiannis, and Chlo{\'e} Clavel. 2024.
\newblock The curious decline of linguistic diversity: Training language models
  on generated text.
\newblock In \emph{Findings of the Association for Computational Linguistics:
  NAACL}.

\bibitem[{Hale(2001)}]{hale2001probabilistic}
John Hale. 2001.
\newblock A probabilistic earley parser as a psycholinguistic model.
\newblock In \emph{Proceedings of NAACL}, pages 159--166.

\bibitem[{Herel and Mikolov(2024)}]{herel2024collapse}
David Herel and Tom{\'a}{\v{s}} Mikolov. 2024.
\newblock Collapse of self-trained language models.
\newblock \emph{arXiv preprint arXiv:2404.02305}.

\bibitem[{Holtzman et~al.(2020)Holtzman, Buys, Du, Forbes, and
  Choi}]{holtzman2020curious}
Ari Holtzman, Jan Buys, Li~Du, Maxwell Forbes, and Yejin Choi. 2020.
\newblock The curious case of neural text degeneration.
\newblock In \emph{International Conference on Learning Representations}.

\bibitem[{Juzek and Ward(2025)}]{juzek2025delve}
Tom Juzek and Adrian Ward. 2025.
\newblock Why does chatgpt ``delve'' so much? exploring the sources of lexical
  overrepresentation in large language models.
\newblock In \emph{COLING}.

\bibitem[{Kobak et~al.(2025)Kobak, Gonz{\'a}lez-M{\'a}rquez, Horv{\'a}t, and
  Lause}]{kobak2025delve}
Dmitry Kobak, Rita Gonz{\'a}lez-M{\'a}rquez, Em{\H{o}}ke-{\'A}gnes Horv{\'a}t,
  and Jan Lause. 2025.
\newblock Delving into llm-assisted writing in biomedical publications through
  excess vocabulary.
\newblock \emph{Science Advances}.

\bibitem[{Levy(2008)}]{levy2008expectation}
Roger Levy. 2008.
\newblock Expectation-based syntactic comprehension.
\newblock \emph{Cognition}, 106(3):1126--1177.

\bibitem[{Liang et~al.(2024)}]{liang2024monitoring}
Weixin Liang and 1 others. 2024.
\newblock Monitoring ai-modified content at scale: A case study on the impact
  of chatgpt on ai conference peer reviews.
\newblock In \emph{International Conference on Machine Learning (ICML)}.

\bibitem[{Mitchell et~al.(2023)Mitchell, Lee, Khazatsky, Manning, and
  Finn}]{mitchell2023detectgpt}
Eric Mitchell, Yoonho Lee, Alexander Khazatsky, Christopher~D Manning, and
  Chelsea Finn. 2023.
\newblock Detectgpt: Zero-shot machine-generated text detection using
  probability curvature.
\newblock In \emph{International Conference on Machine Learning (ICML)}.

\bibitem[{Padmakumar and He(2024)}]{padmakumar2024diversity}
Vishakh Padmakumar and He~He. 2024.
\newblock Does writing with language models reduce content diversity?
\newblock In \emph{International Conference on Learning Representations
  (ICLR)}.

\bibitem[{Peterson and Christiano(2025)}]{knowledgecollapse2025}
Jared~C. Peterson and Paul Christiano. 2025.
\newblock Knowledge collapse in language models.
\newblock \emph{arXiv preprint arXiv:2509.04796}.

\bibitem[{Radford et~al.(2019)Radford, Wu, Child, Luan, Amodei, and
  Sutskever}]{radford2019gpt2}
Alec Radford, Jeffrey Wu, Rewon Child, David Luan, Dario Amodei, and Ilya
  Sutskever. 2019.
\newblock Language models are unsupervised multitask learners.

\bibitem[{Seddik et~al.(2024)Seddik, Chen, Hayou, Youssef, and
  Debbah}]{seddik2024model}
Mohamed El~Amine Seddik, Suei-Wen Chen, Soufiane Hayou, Pierre Youssef, and
  Merouane Debbah. 2024.
\newblock How bad is training on synthetic data? a statistical analysis of
  language model collapse.
\newblock \emph{arXiv preprint arXiv:2404.05090}.

\bibitem[{Shumailov et~al.(2024)Shumailov, Shumaylov, Zhao, Papernot, Anderson,
  and Gal}]{shumailov2024curse}
Ilia Shumailov, Zakhar Shumaylov, Yiren Zhao, Nicolas Papernot, Ross Anderson,
  and Yarin Gal. 2024.
\newblock Ai models collapse when trained on recursively generated data.
\newblock \emph{Nature}, 631:755--759.

\bibitem[{Sourati et~al.(2025)Sourati, Jin, Sch{\"o}lkopf, and
  Sachan}]{sourati2025shrinking}
Zhijing Sourati, Zhijing Jin, Bernhard Sch{\"o}lkopf, and Mrinmaya Sachan.
  2025.
\newblock The shrinking landscape of linguistic diversity.
\newblock \emph{arXiv preprint arXiv:2502.11266}.

\bibitem[{Tercon and Dobrovoljc(2025)}]{tercon2025stylometric}
Andraz Tercon and Kaja Dobrovoljc. 2025.
\newblock Linguistic characteristics of ai-generated text: A survey.
\newblock \emph{arXiv preprint arXiv:2510.05136}.

\bibitem[{Vanmassenhove(2025)}]{vanmassenhove2025losing}
Eva Vanmassenhove. 2025.
\newblock Losing our tail, again: (un)natural selection \& multilingual llms.
\newblock \emph{arXiv preprint arXiv:2507.03933}.

\bibitem[{Welleck et~al.(2020)Welleck, Kulikov, Roller, Dinan, Cho, and
  Weston}]{welleck2020neural}
Sean Welleck, Ilia Kulikov, Stephen Roller, Emily Dinan, Kyunghyun Cho, and
  Jason Weston. 2020.
\newblock Neural text generation with unlikelihood training.
\newblock In \emph{International Conference on Learning Representations}.

\bibitem[{Wu et~al.(2024)}]{wu2024fingerprint}
Yiwei Wu and 1 others. 2024.
\newblock A corpus-based multidimensional analysis of linguistic features
  between human and chatgpt text.
\newblock \emph{Applied Linguistics}.

\bibitem[{Zanotto and Aroyehun(2024)}]{zanotto2024fingerprints}
Francesco Zanotto and Segun~Taofeek Aroyehun. 2024.
\newblock Human variability vs. machine consistency: A comprehensive
  multi-domain analysis of linguistic fingerprints.
\newblock \emph{arXiv preprint arXiv:2412.03025}.

\bibitem[{Zhang et~al.(2022)Zhang, Roller, Goyal, Artetxe, Chen, Chen, Dewan,
  Diab, Li, Lin et~al.}]{zhang2022opt}
Susan Zhang, Stephen Roller, Naman Goyal, Mikel Artetxe, Moya Chen, Shuohui
  Chen, Christopher Dewan, Mona Diab, Xian Li, Xi~Victoria Lin, and 1 others.
  2022.
\newblock Opt: Open pre-trained transformer language models.
\newblock \emph{arXiv preprint arXiv:2205.01068}.

\end{thebibliography}
\end{document}